\documentclass[11pt]{article}

\usepackage[final]{acl}

\usepackage{times}
\usepackage{latexsym}

\usepackage{bbm}

\usepackage[T1]{fontenc}

\usepackage[utf8]{inputenc}

\usepackage{microtype}

\usepackage{inconsolata}

\usepackage{graphicx}

\usepackage{url}
\usepackage{subcaption}
\usepackage{booktabs}
\usepackage{amsmath}
\usepackage{amssymb}

\usepackage{listings}
\usepackage{xcolor}

\usepackage{xurl}

\title{Mediocrity is the key for LLM as a Judge Anchor Selection}

\author{
 \textbf{Shachar Don-Yehiya\textsuperscript{1,2}} \qquad
 \textbf{Asaf Yehudai \textsuperscript{1,2}} \qquad
 \textbf{Leshem Choshen \textsuperscript{2,3,4}} \qquad
 \textbf{Omri Abend\textsuperscript{1}} \\
 \textsuperscript{1}The Hebrew University of Jerusalem,
 \textsuperscript{2}IBM Research,
 \textsuperscript{3}MIT,
 \textsuperscript{4}MIT-IBM Watson AI Lab \\
   \texttt{\{first.last\}@mail.huji.ac.il}
}

\begin{document}
\maketitle

\begin{abstract}
The ``LLM-as-a-judge'' paradigm has become a standard method for evaluating open-ended generation. To address the quadratic scalability costs of pairwise comparisons, popular benchmarks like Arena-Hard and AlpacaEval compare all models against a single anchor. However, despite its widespread use, the impact of anchor selection on the reliability of the results remains largely unexplored. In this work, we systematically investigate the effect of anchor selection by evaluating 22 different anchors on the Arena-Hard-v2.0 dataset. We find that the choice of anchor is critical: a poor anchor can dramatically reduce correlation with human rankings. 
We identify that common anchor choices (best-performing and worst-performing models) make poor anchors. Because these extreme anchors are consistently better or worse than all other models, they are seldom indicative of the relative ranking of the models. 
We further quantify the effect size of anchor selection, showing it is comparable to the selection of a judge model.
We conclude with actionable recommendations. First, we conduct a power analysis, and compute sufficient benchmark sizes for anchor-based evaluation, finding that standard benchmark sizes are insufficient for pairwise evaluation and fail to distinguish between competitive models reliably.
Second, we provide guidelines for selecting informative anchors to ensure reliable and efficient evaluation practices.
\end{abstract}


\section{Introduction}\label{sec:intro}



Traditional reference-based metrics \citep{papineni-etal-2002-bleu, lin-2004-rouge} are often ill-suited for the open-ended nature of modern LLM applications \citep{liu-etal-2016-evaluate}. Consequently, "LLM as a Judge" (LMJ)—using one model to evaluate another—has emerged as a scalable alternative that correlates highly with human evaluation \citep{chiang-lee-2023-large, liu-etal-2023-g}, despite some potential for bias \citep{wang-etal-2024, saito2023verbosity}.

\begin{figure}[t]
\centering
\includegraphics[width=\columnwidth]{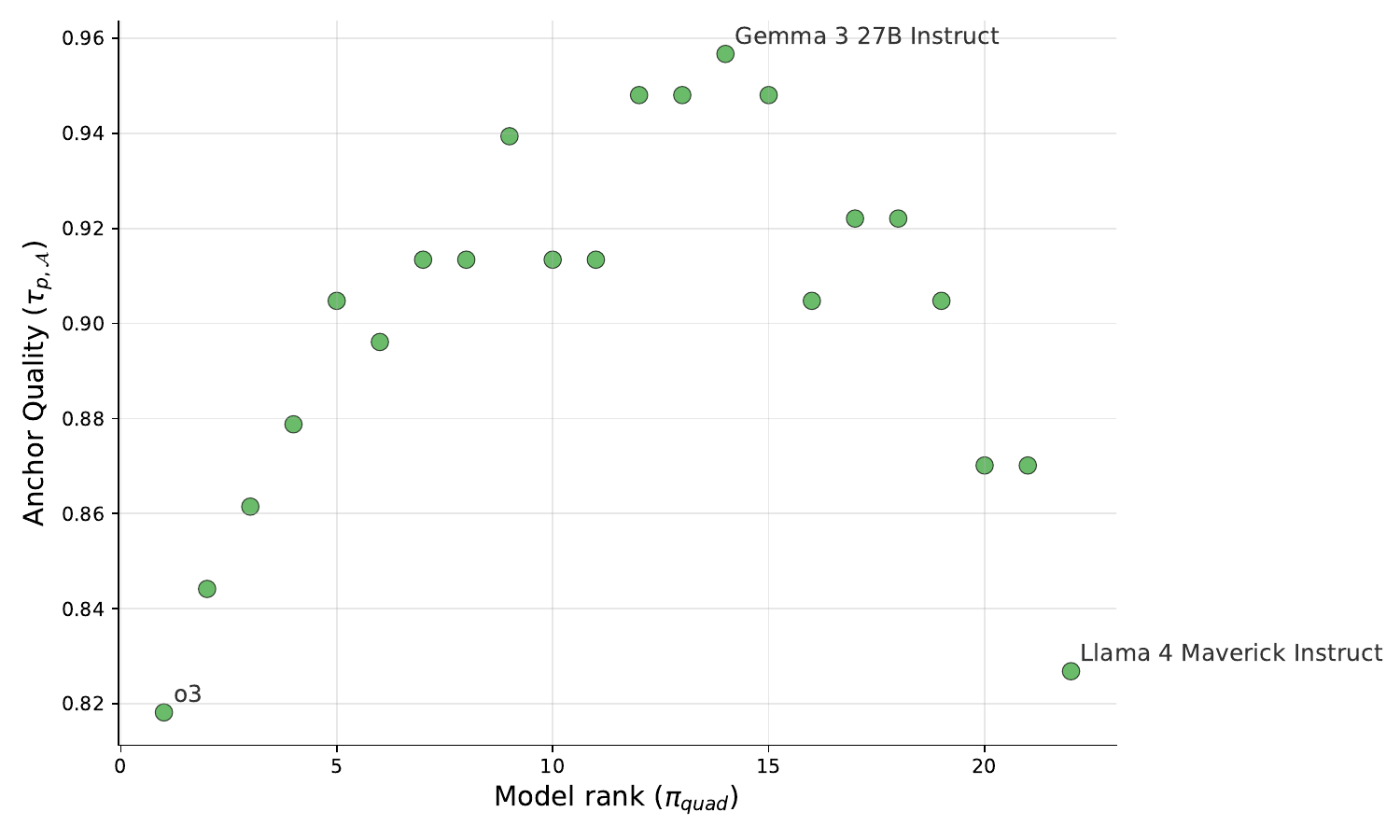}
\caption{Kendall’s $\tau$ correlation ($\tau_{p, \mathcal{A}}$) plotted against anchor position. The y-axis shows the correlation between the anchor-based ranking and the quadratic ranking $\pi_{quad}$, while the x-axis represents the anchor's position (rank) in $\pi_{quad}$. This reveals an inverted U-shaped relationship: top and bottom-ranked models correlate poorly with the gold standard, making them suboptimal anchors. The judge is \texttt{Deepseek-v3}.}
\label{fig:arch_deepseek_vs}
\end{figure}

A primary setting for LMJ is pairwise comparisons. Given an instruction, the judge compares the responses of two models and states a preference (e.g., the first is better).
The main drawback of this approach is that the cost of evaluation grows fast. Specifically, as the number of evaluated models increases, the number of model pairs to compare grows quadratically \citep{Evaluating_llmj}.

To provide better scalability, a common practice is to select an {\it anchor} model and compare all other models to it.
This practice was adopted by popular benchmarks such as the Arena-Hard \citep{li2024crowdsourceddata} and AlpacaEval \citep{alpaca_eval} evaluation frameworks, and is therefore used by many \citep[e.g.,][]{pombal2025zeroshotbenchmarkingframeworkflexible, dubois2024length, raju-etal-2024-constructing, gera-etal-2025-justrank, donyehiya2025naturallyoccurringfeedbackcommon, dpo, kto, meng2024simpo, hong-etal-2024-orpo, vicuna2023, tang2025beyond, chen2025compo, xu2025magpie}. 

Although frequently done, few works study the effects of anchor-based evaluation, mostly focusing on the validity of the transitivity assumption \citep{Non_Transitivity, wang2025trustjudge}, that states that if models $A,B,Anchor$ show $A<Anchor$ and $Anchor<B$, then $A<B$. They demonstrate that this assumption does not hold, and as an alternative suggest dynamic matching strategies \citep{liusie-etal-2024-efficient, son2025arena}, complicating the evaluation process. 

Rather than suggesting alternatives, we study the best practices of using anchors.
We start by empirically examining the effect of anchor choice. We conduct a large-scale analysis involving over $850K$ pairwise comparisons across 22 different anchors on the Arena-Hard-v2.0 dataset. We find that a bad anchor can lead to up to \textbf{$.30/.19$} drop in correlation with human/quadratic rankings. 
Notably, we observe an inverted U-shaped relationship between model capability and anchor quality: top-performing (`strong') and low-performing (`weak') models make the worst anchors (the tails of the U), while `mediocre' models provide the highest correlation (see Fig.~\ref{fig:arch_deepseek_vs}). This is in sharp contrast to the common practice of using strong or weak models as anchors, as they provide simple `baseline' or `gold' standards for comparison \citep{li2024crowdsourceddata, xu2025magpie}.

To better understand the last phenomenon, we look at the win-rate distributions of different anchors (\S\ref{sec:win_rate}). We see that overly strong/weak anchors induce skewed distributions. Our results show that these are less helpful, as many of the samples are less informative. For example, \textit{o3} wins against all the other models in about $500/750$ of the benchmark's samples, wasting $2/3$ of the evaluation budget. 

To further examine the statistical implication of this observation, we run a power analysis that takes into account the `informativeness rate' of the comparisons against the anchor (\S\ref{sec:informative}). We find that for a small effect size ($+5\%$) and an average informativeness rate, the estimated number of samples is larger than the size of the Arena-Hard-v2.0 dataset. This indicates that the current benchmark is statistically insufficient to reliably distinguish between competitive models in an anchor-based setting. 

Finally, we broaden our analysis to practical mitigation strategies and the relative importance of the anchor. We vary the dataset size (\S\ref{sec:data_size}) to test the robustness of our findings, and examine the effect of using multiple anchors (\S\ref{sec:multi_anchors}). Crucially, we compare the impact of selecting an anchor against the impact of selecting a judge model. We conclude that choosing an anchor is a critical factor in evaluation reliability, comparable in its effect to choosing a judge (\S\ref{sec:vs_judges}). 

Based on these insights, we suggest a decision framework for pairwise evaluation (Fig.~\ref{fig:tree}). We advise avoiding anchor-based evaluation when possible—specifically for small model sets ($N \le 3$) or when a natural baseline exists. For leaderboard settings where anchors are unavoidable, we recommend selecting "mediocre" rather than state-of-the-art models to maximize statistical power, and explicitly reporting the anchor's informativeness to ensure validity.

We release about $900K$ judgments generated for this study\footnote{Code: \url{https://github.com/IBM/Anchor-Selection}, Data: \url{https://huggingface.co/datasets/ibm-research/900K-Judgements}.}.


\section{Task Formulation} \label{sec:task_def}

In this work, we study the use of LLM-based judges for determining the relative quality of systems over a given set of user instructions. Henceforth, \textit{System} or \textit{Model} refers to an LLM that performs a task, and \textit{Judge} refers to the LLM that compares the quality of such systems. Specifically, we focus on the \textit{pairwise anchor-based} evaluation setting, and assume that the transitivity assumption (\S\ref{sec:intro}) holds at least to some extent.

Formally, we begin with a set of $M$ systems $\mathcal{M} = \{m_j\}_{j=1}^{M}$, and $N$ user instructions $\mathcal{I} = \{x_i\}_{i=1}^{N}$. 
Each system produces a response for each instruction, denoted with
$\mathcal{R} = \{r_{i,j}\}_{i=1, j=1}^{N, M}$, 
such that $m_j(x_i) = r_{i,j}$.

In the anchor-based setting, one system is designated as the \textit{anchor}, denoted $m_{\mathcal{A}} \in \mathcal{M}$. A judge from $\mathcal{J} = \{J_p\}_{p=1}^{P}$ is tasked with comparing a target response $r_{i,j}$ against the anchor's response $r_{i,\mathcal{A}}$ for the same instruction $x_i$. 

The judge maps a triplet of instructions and two candidate responses to a preference verdict:
$$J_p(x_i, r_{i,j}, r_{i,\mathcal{A}}) = v_{i,j}^{p,\mathcal{A}} \in \{-2,-1, 0, 1, 2\}$$
where $2/1$ represents a clear/slight win for the target model $m_j$ over the anchor model $m_{\mathcal{A}}$, $-2/-1$ represent a loss, and $0$ a tie.
Once a judge $J_p$ evaluates all systems against the chosen anchor $m_{\mathcal{A}}$, we obtain a verdict matrix $V^{p, \mathcal{A}} \in \mathbb{R}^{N\times M}$.

In order to quantify system-level quality, we apply an \textit{aggregation method}. The aggregation method maps the verdict data to a system-level score vector $\mathbf{s} \in \mathbb{R}^{M}$. 
We consider two aggregation methods commonly used in anchor-based evaluation:

\begin{itemize}
    \item \textbf{Win-Rate:} 
    We collapse the verdicts into $\{0, 0.5, 1\}$ and compute the average win-rate against the anchor: $s_j = \frac{1}{N} \sum_{i=1}^N v_{i,j}^{p, \mathcal{A}}$. To score the anchor itself, we compute its average win-rate: $s_\mathcal{A}=1 - \frac{\sum_{j\neq \mathcal{A}}{s_j}}{M - 1}$.
    \item \textbf{Bradley-Terry (BT):} We collapse the verdicts into $\{-1, 0, -1\}$ and follow \citet{chatbot_arena} estimating the vector of BT coefficients\footnote{In the Chatbot Arena notebook (\url{https://colab.research.google.com/drive/1KdwokPjirkTmpO_P1WByFNFiqxWQquwH}), they demonstrated that the Elo score \citep{elo1967proposed} is noisy for model ranking, as it is highly influenced by the battles order \citep{boubdir-etal-2023-elo}. To obtain more stable results, they used Bradley-Terry \citep{Bradley1952RANKAO}.} $s_j$ that maximizes the likelihood of the observed pairwise verdicts in $V^{p, \mathcal{A}}$. This model posits that the probability of system $i$ beating system $j$ is $P(i \succ j) = \frac{e^{s_i}}{e^{s_i} + e^{s_j}}$.
\end{itemize}

Ordering the scores in $\mathbf{s}$ induces a ranking over the system set $\mathcal{M}$, denoted with $\pi(\mathbf{s})$. We evaluate the judge $J_p$ with anchor $m_{\mathcal{A}}$ by comparing this induced ranking against a golden ranking $\pi^*$ derived from quadratic comparisons or human annotations. Specifically, we define the \textit{anchor quality} to be the Kendall's $\tau$ correlation coefficient:
\begin{equation}
    \tau_{p, \mathcal{A}} = \text{Kendall}(\pi(\mathbf{s}), \pi^*)
\end{equation}


\section{Experimental Setup} \label{sec:setup}

\paragraph{Data.}
We use the Arena-Hard-v2.0 benchmark that contains 500 challenging real-world user queries (open-ended software engineering problems, math questions, logic puzzles, etc.) and 250 creative writing queries sourced from Chatbot Arena \citep{chatbot_arena}.
We replicate the results for the AlpacaEval dataset, which includes $805$ instructions from the test sets of Self-instruct \citep{wang-etal-2023-self-instruct}, Open Assistant\footnote{https://github.com/LAION-AI/Open-Assistant}, Anthropic’s HH-RLHF \citep{bai2022training}, Vicuna \citep{Evaluating_llmj, vicuna2023}, and Koala \citep{koala_blogpost_2023}.

\paragraph{Models.}
We examine all models that appear in the Arena-Hard-Auto repository \footnote{\url{https://github.com/lmarena/arena-hard-auto/tree/main}}, and are available in the Chatbot Arena leaderboard (see \S\ref{sec:human_ranking}). Thus, we will be able to compare the automatically extracted ranking to the arena's human ranking. 
We end up with $22$ contemporary models, see App.~\ref{app:full_results} for the full list.
These models are used both as anchors and as competitors.

\paragraph{Judges.}\label{sec:judges}
We experiment with $5$ different judges: \textit{Deepseek-v3} \citep{guo2025deepseek}, \textit{GPT-OSS 120B}, \textit{GPT-OSS 20B} \citep{openai2025gptoss120bgptoss20bmodel}, \textit{Qwen3 235B-A22B Instruct}, and \textit{Qwen3 8B} \citep{qwen3}. We chose the first four models based on their high performance, and the last one as a smaller (and hence cheaper) alternative. We run the judges with their default parameters and use the evaluation prompt from the Arena-Hard-Auto repository. 


\subsection{Extracting Anchor-Based Ranking}\label{sec:anchor_based}

Given a judge $J_p$ and an anchor $m_{\mathcal{A}}$, 
we present the judge with a user query $x_i$ and two model responses $(r_{i,\mathcal{A}}, r_{i,j})$, one generated by the anchor and one by another model $m_j$. We then parse the judge's output to extract its verdict $v_{i,j}^{p,\mathcal{A}}$. Repeating this for all the benchmark samples and for the $22$ evaluated models, we end up with $750 \cdot 22 = 16,500$ comparisons per anchor and judge.
We use these comparisons to calculate the model's win-rates against the anchor to extract a ranking $\tau_{p, \mathcal{A}}$ (see \S\ref{sec:task_def}).

\subsection{Extracting Quadratic (\textit{Gold}) Ranking}\label{sec:gold}

We run the `quadratic' comparisons, i.e., for each instance of the benchmark, we compare the responses of all possible pairs of models.
This sums up to $\binom{22}{2} \times 750 = 173,250$ comparisons per judge, and $173,250 \cdot 5 = 866,250$ in total. The comparisons of a judge can be summarized into a $22 \times 22$ win-rate matrix, $V^{p, \mathcal{A}}$.
As the anchor-based ranking is an approximation of the quadratic ranking, we refer to the quadratic ranking as our `gold'.
Given the win-rate matrix, we use BT to extract the `quadratic ranking', $\pi_{quad}$.
To complete the picture, we obtain a human ranking as well, see the next section.

\subsection{Human Ranking}\label{sec:human_ranking}
To obtain a human ranking $\pi_{human}$, we use the model's scores from the Chatbot Arena text leaderboard.\footnote{\url{https://lmarena.ai/leaderboard/text}} Chatbot Arena collects human-annotated battles between pairs of models' responses and then aggregates the battles with Bradley-Terry into model scores and a continuously updating leaderboard. As the battles data were not available, we took the aggregated scores.

\begin{table}[t]
\centering
\small
\begin{tabular}{p{0.55\linewidth} c c}
\hline
\textbf{Anchor} & $\tau_{\text{quad}}$ & $\tau_{\text{human}}$ \\
\hline
Gemma 3 27B Instruct & .957 & .514 \\
Qwen3 30B A3B & .948 & .495 \\
o1 & .948 & .476 \\
o3 Mini & .948 & .486 \\
Claude 3.7 Sonnet thinking 16k & .939 & .467 \\
Athene V2 Chat & .922 & .429 \\
Claude 3.5 Sonnet & .922 & .486 \\
o3 Mini High & .913 & .476 \\
GPT-4.5 (Preview) & .913 & .495 \\
QwQ 32B & .913 & .533 \\
GPT-4.1 & .913 & .457 \\
GPT-4.1 Mini & .905 & .543 \\
GPT-4.1 Nano & .905 & .505 \\
Qwen3 32B & .905 & .581 \\
o4 Mini & .896 & .448 \\
DeepSeek-R1 & .879 & .486 \\
Llama 3.1 Nemotron 70B Instruct & .870 & .476 \\
Qwen2.5 72B Instruct & .870 & .381 \\
Gemini 2.5 Flash & .861 & .476 \\
Qwen3 235B A22B & .844 & .571 \\
Llama 4 Maverick Instruct & .827 & .448 \\
o3 & .818 & .324 \\
\hline
\textit{Average correlation:} & .901 & .480 \\
\textit{Standard deviation:} & .039 & .057 \\
\hline
\end{tabular}
\caption{Kendall's tau correlations of each anchor-based ranking with quadratic and human rankings.}\label{tab:corr_combined}
\end{table}

\section{Results}

For each judge $J_p$, we measure its quality by Kendall's $\tau$ correlation, $\tau_{p, \mathcal{A}}$, between the anchor-based ranking induced by $m_\mathcal{A}$ (\S\ref{sec:anchor_based}) and the quadratic ranking $\pi_{quad}$ (\S\ref{sec:gold}). Table \ref{tab:corr_combined} shows the results for \texttt{deepseek-v3} as the judge ($J_p$). The tables for the other four judges are found in App.~\ref{app:full_results}.
Although all correlations are significant with $p<0.05$, we observe that anchors vary in their quality, as shown by a $.14$ gap in $\tau_{p, \mathcal{A}}$ between the best and worst anchor choices.

We repeat the analysis now comparing to the human ranking, $\pi_{\text{human}}$ (\S\ref{sec:human_ranking}), as a reference. We observe similar trends, but with an even larger sensitivity, showing a drop of $.19$ in $\tau_{p, \mathcal{A}}$ between the best and worst anchor choices.

Crucially, the identity of the \textit{worst} anchor is consistent across both the quadratic and human rankings: the \textit{o3} model. Note that this is also the \textit{top-performing} model in our set. In what follows, we study this relation between an anchor's performance, and its effectiveness as a reference point.

\subsection{Correlation with Model Ranking} \label{sec:corr}

To examine the relation between an anchor's performance and its effectiveness as an anchor, 
Fig.~\ref{fig:arch_deepseek_vs} plots the correlation $\tau_{p, \mathcal{A}}$ against the rank of the anchor $m_\mathcal{A}$ within the quadratic ranking $\pi_{quad}$ with \textit{Deepeseek V3} as the judge (the full plot with all labels is provided in App.~\ref{app:full_results}). The plot reveals a distinct inverted U-shape, where anchors at the edges of $\pi_{quad}$ (the top and bottom performing models) consistently yield the lowest performance. This finding challenges common practices, where extreme models are frequently selected as $m_\mathcal{A}$ under the assumption that they provide a strong baseline or a reliable lower-bound \citep[e.g.,][]{li2024crowdsourceddata}. We observe this same inverted U-shape pattern when comparing to the human ranking $\pi_{\text{human}}$ as the ground truth.
Additionally, we replicate the experiment for the other judges 
and on the AlpacaEval dataset; see App.~\ref{app:full_results}.

\begin{figure*}[t]
    \centering
    \begin{subfigure}{0.32\textwidth}
        \centering
        \includegraphics[width=\linewidth]{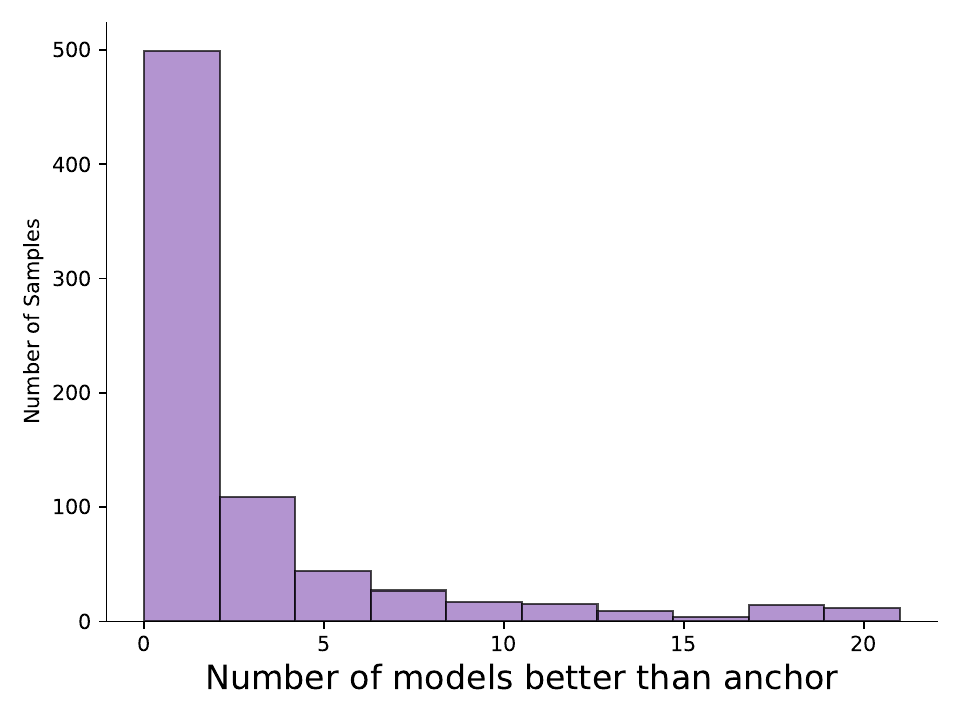}
        \caption{o3}
        \label{fig:o3_dist}
    \end{subfigure}
    \hfill
    \begin{subfigure}{0.32\textwidth}
        \centering
        \includegraphics[width=\linewidth]{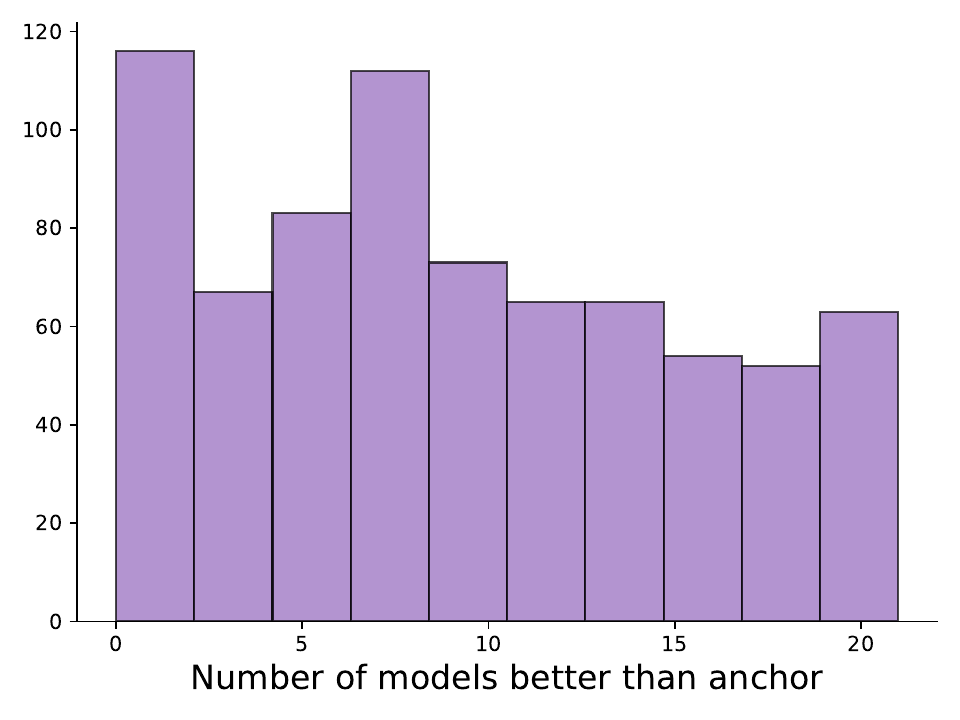}
        \caption{Gemma 3 27B-Instruct}
        \label{fig:gemma_dist}
    \end{subfigure}
    \hfill
    \begin{subfigure}{0.32\textwidth}
        \centering
        \includegraphics[width=\linewidth]{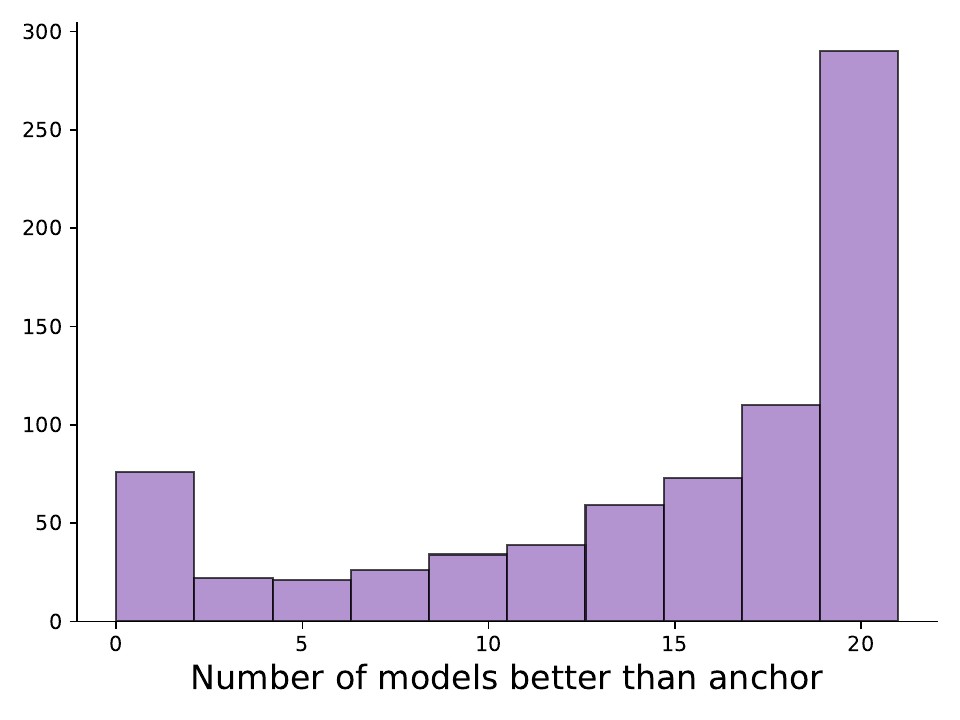}
        \caption{Llama 4 Maverick Instruct}
        \label{fig:llama_dist}
    \end{subfigure}
    
    \caption{Histograms of the frequency of samples (Y-axis) grouped by the number of models that outperformed the anchor (X-axis). A value of 0 on the X-axis indicates samples where the anchor was superior to all other models, while higher values indicate samples where the anchor was frequently outperformed. \textit{o3} (\ref{fig:o3_dist}) shows a positive skew, as most of the data points are clustered on the left, in accordance with \textit{o3} being a strong model that usually beats its opponents. Respectively, we get a negative skew for the low performing \textit{Llama 4 Maverick Instruct} (\ref{fig:llama_dist}). For \textit{Gemma 3 27B-Instruct} (\ref{fig:gemma_dist}) we get a more evenly spread “flatter” distribution.}
    \label{fig:anchor_dist}
\end{figure*}

\subsection{Win-Rate Distribution} \label{sec:win_rate}
To explain the inverted U-shape finding, we examine the win-rate distributions of the anchors against all other models. 
Given an anchor $m_{\mathcal{A}}$, 
for each instruction $x_i$, we count the number of models that win over the anchor
$\sum_{j=1}^{j=M}\mathbbm{1}_{v_{i,j}^{p,\mathcal{A}}>0}$.

Fig.~\ref{fig:anchor_dist} visualizes this for three representative anchors: the top-performing \textit{o3} (right tail), the low-performing \textit{Llama4 Maverick Instruct} (left tail), and the mid-level \textit{Gemma 3 27B instruct} (peak).
For the strong \textit{o3}, we observe a heavy positive skew; data points cluster on the left as the anchor dominates most comparisons. Conversely, we see the opposite negative skew for the weak \textit{Llama4 Maverick Instruct}. However, for the mid-level \textit{Gemma 3 27B instruct}, we observe a flatter, more evenly spread distribution.

This distribution shape directly explains the anchor quality. The strongly skewed distributions at the tails are less informative because they suffer from signal saturation; a significant portion of the samples cannot distinguish between different models. For instance, \textit{o3} defeats all opposing models in roughly $500/750$ of the benchmark's samples. This effectively wastes 2/3 of the evaluation budget, as these comparisons yield no information about the relative strength of the opponents.

\begin{figure}[t]
\centering
\includegraphics[width=\columnwidth]{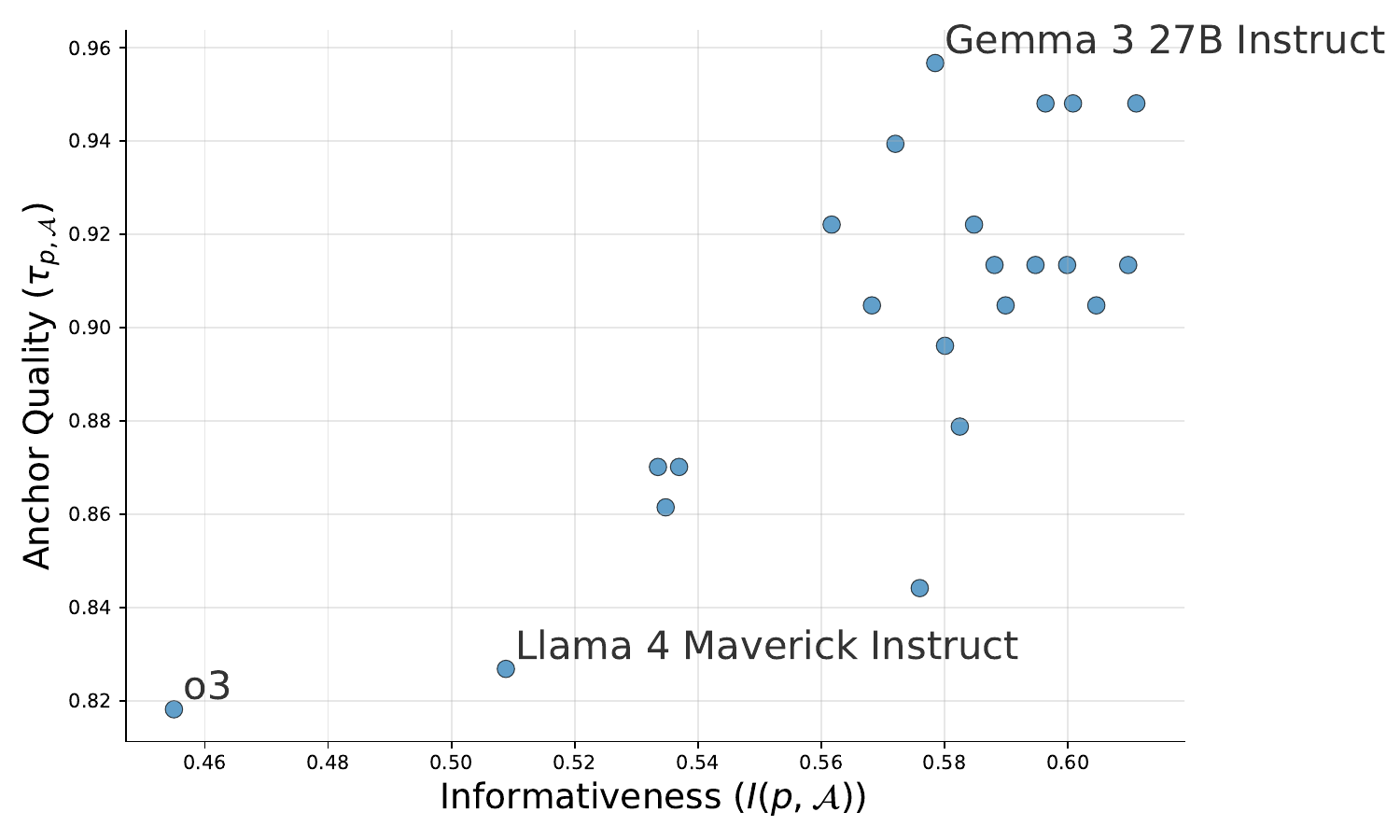}
\caption{Kendall’s $\tau$ correlation ($\tau_{p, \mathcal{A}}$) plotted against anchor informativeness. The y-axis shows the correlation between the anchor-based ranking and the quadratic ranking $\pi_{quad}$, while the x-axis represents the anchor's informativeness $I(p, \mathcal{A)}$. The plot exhibits a positive correlation between anchor quality and anchor informativeness. The judge is \textit{Deepseek-v3}.} \label{fig:informative_deepseek_vs}
\end{figure}

\begin{table}[htbp]
    \captionsetup{justification=raggedright, singlelinecheck=false}

    \resizebox{\columnwidth}{!}{%
    \begin{tabular}{lcccc}
        \toprule
        & & \multicolumn{3}{c}{\textbf{Required Sample Size ($N$)}} \\
        \cmidrule(lr){3-5}
        \textbf{Win Rate} & \textbf{Edge} & \textbf{Base} & \textbf{Total} & \textbf{Total} \\
        (Discordant) & & (No Ties) & (39\% Ties) & (55\% Ties) \\
        \midrule
        55\% & +5\% & 617 & 1,012 & 1,372 \\
        60\% & +10\% & 153 & 251 & 341 \\
        65\% & +15\% & 67 & 110 & 149 \\
        70\% & +20\% & 37 & 61 & 83 \\
        75\% & +25\% & 23 & 38 & 52 \\
        \bottomrule
    \end{tabular}%
    }
    \par\medskip
        \caption{Required Total Sample Sizes (One-Sided Test) adjusted for informativeness rates. \textbf{Base} is the discordant pairs needed for statistical significance using a \textbf{One-Sided Test} ($\alpha=0.05$, Power=0.80). \textbf{Total} columns account for data loss due to ties.}
    \label{tab:power_analysis}
\end{table}

\subsection{Informative Samples} \label{sec:informative}

To quantify the impact on the evaluation quality of the win-rate distributions observed in \S\ref{sec:win_rate}, we measure the prevalence of `informative samples' for each anchor.
For a sample $i$ to be informative for two models $a$ and $b$, their verdicts w.r.t. $\mathcal{A}$, should hold $v_{i,a}^{p,\mathcal{A}} \neq v_{i,b}^{p,\mathcal{A}}$.
We will therefore define the informativeness of an anchor as
$$I(p, \mathcal{A}) = \frac{1}{N\cdot \binom{M}{2}}\sum_{i=1}^N \sum_{a,b\in \mathcal{M}} \mathbbm{1}_{v_{i,a}^{p,\mathcal{A}} \neq v_{i,b}^{p,\mathcal{A}}}$$

Our empirical results reinforce the inverted U-shaped hypothesis. 
The top-performing anchor \textit{o3} yields only $45\%$ informative samples—meaning $55\%$ of the compute budget provides no discriminative signal. In contrast, with the highest informativeness we have \textit{o3 Mini} with $61\%$ informative samples. However, this implies that even in the best scenario roughly $39\%$ of the evaluation budget is inevitably wasted.
See App.~\ref{app:informativeness} for the full results.

To contextualize these rates, note that for the case of verdicts $v_{i,j}^{p,\mathcal{A}} \in \{-1, 0,-1\}$ (no magnitude), and assuming that the transitivity assumption holds, then $I(p, \mathcal{A}) \leq 0.5$ with equity when the anchor is ranked exactly in the middle. That is, the anchor-based setting inherently limits the informativeness.   

Table~\ref{tab:power_analysis} 
shows the number of samples ($N$) needed to achieve statistical significance (Power=80\%, $\alpha=0.05$) in a sign test
between two models. The null hypothesis ($H_0$) is that model B is better than or equal to model A according to the judge’s predictions.
We can see that for a small effect size, we will need $617$ samples. However, these samples should be \textit{informative}, as the sign test ignores the tied cases. Thus, we will need $$N_{total}=\frac{N}{I(p, \mathcal{A})}$$ samples\footnote{Note that this is an approximation, as $I(p, \mathcal{A})$ is averaged across all model pairs, whereas $N$ varies with effect size.}, and in the case of \textit{o3} we will have $N_{total}=1372$, far more than the $750$ samples of the dataset.

To make better use of the judgments, we can employ a weighted approach like the Wilcoxon signed-rank test. Instead of collapsing the results into three options (tie, model A wins, model B wins), Wilcoxon takes into account the margin of the win (i.e., model A is slightly better than the anchor while model B is in a tie with it $\neq$ model A is strongly better than the anchor while model B is strongly worse than the anchor). As this test has stronger assumptions about the data, we run a simulation to find $N$, see App.\ref{app:power_analysis}. We find that for a small effect size of $+5\%$ and an average of $54\%$ informative samples (the mean informativeness we have for our empirical distributions), we will need $N=930$, about $200$ samples less than the sign test ($1143$), but still more than the dataset size.

Finally, we provide empirical support for the link between sample efficiency and ranking accuracy. Fig.~\ref{fig:informative_deepseek_vs} plots the correlation of each anchor's resulting ranking with the quadratic ranking against its informativeness (for the full plot with all labels see App.~\ref{app:informativeness}). We observe a strong positive relationship ($R^2=0.5940$): as the informativeness increases, the anchor-based ranking aligns more closely with the quadratic ranking. This suggests that maximizing the number of informative samples is not solely a question of computational efficiency, but also yields more reliable results.

\section{Robustness and Sensitivity Analysis}

Having identified the roles of win-rate distributions and informative samples in anchor-based evaluation, we now examine how the evaluation pipeline responds to different settings. We start by testing whether increasing the scale of the evaluation (increasing the dataset size or adding more anchors, see App.~\ref{sec:multi_anchors}) can mitigate the performance drop of anchor-based evaluation vs. quadratic evaluation. 
We then compare the magnitude of the effect of selecting an anchor in comparison to that of a judge model (finding them to be comparable).
Finally, we estimate the anchor informativeness with fewer samples to allow an informed anchor selection before running the evaluation on the full dataset.

\begin{figure}[t]
    \centering
    \includegraphics[width=\columnwidth]{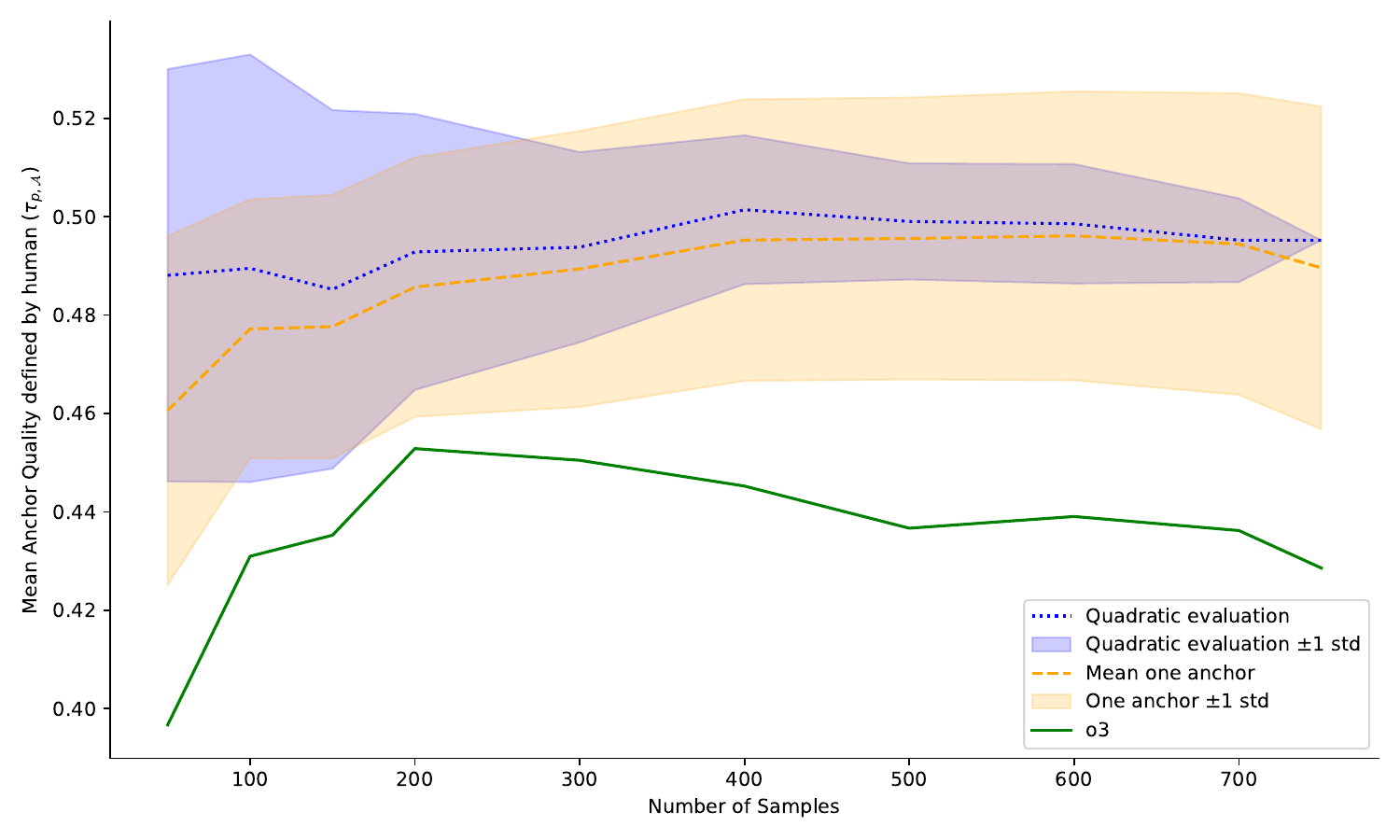}
    \caption{Mean $\tau_{p, \mathcal{A}}$ with respect to human ranking averaged over random sample selections as a function of sample size. 
    As the number of samples grows, the variance of the quadratic evaluation correlation decreases. Simultaneously, the mean anchor-based correlation improves, eventually converging with the quadratic correlation at approximately $600$ samples. This is not the case for each particular anchor choice, see \textit{o3} correlation. This demonstrates that anchor-based ranking is more affected by the dataset size than the quadratic ranking. The judge is \texttt{Deepseek-v3}.}
    \label{fig:num_samples}
\end{figure}

\subsection{Number of Samples} \label{sec:data_size}
We next investigate the sample efficiency of anchor-based methods compared to the quadratic approach.
We varied the number of instructions $N$, sampling 10 sets of samples of sizes $50$ to $750$.
We run BT to aggregate the quadratic ranking. We do the same for each anchor $\mathcal{A}$, extracting their anchor-based ranking and their correlation with the human ranking $\tau_{p, \mathcal{A}}$. We repeat the process $30$ times and average over the resulting correlations.

Our results for \textit{Deepseek-V3} as the judge (Fig.~\ref{fig:num_samples}) present a distinct difference in stability. 
Although the standard deviation of the quadratic correlation shrinks as the number of samples grows, the mean correlation does not change much. In contrast, anchor-based rankings are highly sensitive to dataset size. 
The mean correlation of the anchor-based rankings across all the anchors improves to the point that the quadratic correlation and the mean anchor-based correlation are pretty close (around $600$ samples).
Note that this is not the case for each particular anchor choice, as the \textit{o3} correlation remains far below even when the number of samples grows. Results for the other judges are provided in App.~\ref{app:num_samples}.

We conclude that anchor-based ranking is more affected by the size of the dataset than the quadratic evaluation. This is in line with the results from \S\ref{sec:informative}, where we saw that in anchor-based evaluation a significant portion of the dataset is wasted (up to $55\%$ of the comparisons are not informative) and therefore the effective dataset is smaller.

\begin{table}[htbp]
    \resizebox{\linewidth}{!}{%
        \begin{tabular}{lccccccc}
            \toprule
             & \textbf{Quadratic} & \multicolumn{3}{c}{\textbf{Quadratic vs}} & \multicolumn{3}{c}{\textbf{Human vs}} \\
             \cmidrule(lr){3-5} \cmidrule(lr){6-8}
            \textbf{Judge} & \textbf{vs Human} & \textbf{Best} & \textbf{Worst} & \textbf{$\Delta$} & \textbf{Best} & \textbf{Worst} & \textbf{$\Delta$} \\
            \midrule
            DeepSeek-V3 & .495 & .983 & .844 & .139 & .562 & .371 & .191 \\
            GPT-OSS 120B & .514 & .957 & .766 & .191 & .581 & .371 & .210 \\
            GPT-OSS 20B & .419 & .974 & .784 & .190 & .505 & .200 & .305 \\
            Qwen3 235B A22B& .429 & .965 & .818 & .147 & .505 & .276 & .229 \\
            Qwen3 8B & .333 & .957 & .827 & .130 & .410 & .229 & .181 \\
            \bottomrule
        \end{tabular}%
    }
    \medskip
    \caption{Kendall's $\tau$ correlation coefficients comparing Quadratic/Human and Anchor-based ranking. `Best' refers to the best anchor choice, i.e., the anchor that yields the ranking with the highest correlation to the quadratic. `Worst' is the anchor that yields the ranking with the lowest correlation. The $\Delta$ columns show the gain from the best vs. the worst anchor choice.} 
    \label{tab:judges}
\end{table}

\subsection{Comparing the Effect of Anchor vs. Judge Selection} \label{sec:vs_judges}

Finally, we contextualize the magnitude of the anchor effect by comparing it to the judge effect. Typically, great effort is spent selecting the strongest judge model \citep{tan2025judgebench, thakur-etal-2025-judging}.
This subsection addresses the question of whether selecting an anchor has a similar effect (and is therefore equally important).


Table~\ref{tab:judges} summarizes the results (full results are provided in App.~\ref{app:full_results}). We see that the anchor choice has a comparable or often larger impact on performance than the choice of judge model. The influence of the judge is clearly visible in the "Quadratic vs Human" column, where the difference in the correlation with human judgments between the worst and best judges is $1.81$. The impact of the anchor choice is even more pronounced: the difference between the worst and best anchors (in terms of correlation with human judgments) is $0.181-0.305$, depending on the judge. 

Moreover, the effect of selecting a good anchor seems to be complementary to the choice of the judge. Indeed, similar anchor effects are presented for judges of different performance levels.

\subsection{Estimating Anchor Informativeness}\label{sec:estimate}

Our experiments showed that the accuracy of the anchor-based ranking is tightly linked to the percentage of informative samples (\S\ref{sec:informative}).
Thus, we would like to choose an anchor that has a high percentage of informative samples. 
Running the full evaluation for one anchor requires generating $N \cdot (M + 1) $ model responses, and $N \cdot M$ judgments.
We would like to estimate the anchor informativeness with fewer judgments.

To validate this estimation strategy, we conduct an experiment where we varied the number of evaluated models $M$ within the range $[3, 22]$. For each $M$, we randomly select $M$ models and $10$ samples from the dataset. We then calculate the informativeness rate for each of the $22$ potential anchors and measure the Pearson correlation against the rates derived from the full dataset, repeating the process $30$ times. 

The results indicate a strong predictive capability: for $M \geq 3$, the Pearson correlation is above $0.86$, and for $M \geq 8$, above $0.91$. Additionally, across all values of $M$, the estimation successfully identified the least informative anchors—specifically the best and worst-performing models—consistently placing them in the bottom three rankings.

In terms of absolute scores, in \S\ref{sec:informative} we saw that the empirical informativeness rate ranges in $(0.44,0.61)$. Here with $10$ samples only, we see similar trends, with a maximum informativeness rate of $0.65$ (\textit{GPT-4.5 (Preview)} and \textit{o3 Mini}) and a minimum of $0.42$ (\textit{o3}).
These findings confirm that good anchors can be reliably identified with fewer samples, which is of practical importance, where the more informative anchors (the mid-performing ones) are not known in advance. For the full results, see App.~\ref{app:estimate}.


\begin{figure}[t]
    \centering
    \includegraphics[width=\columnwidth]{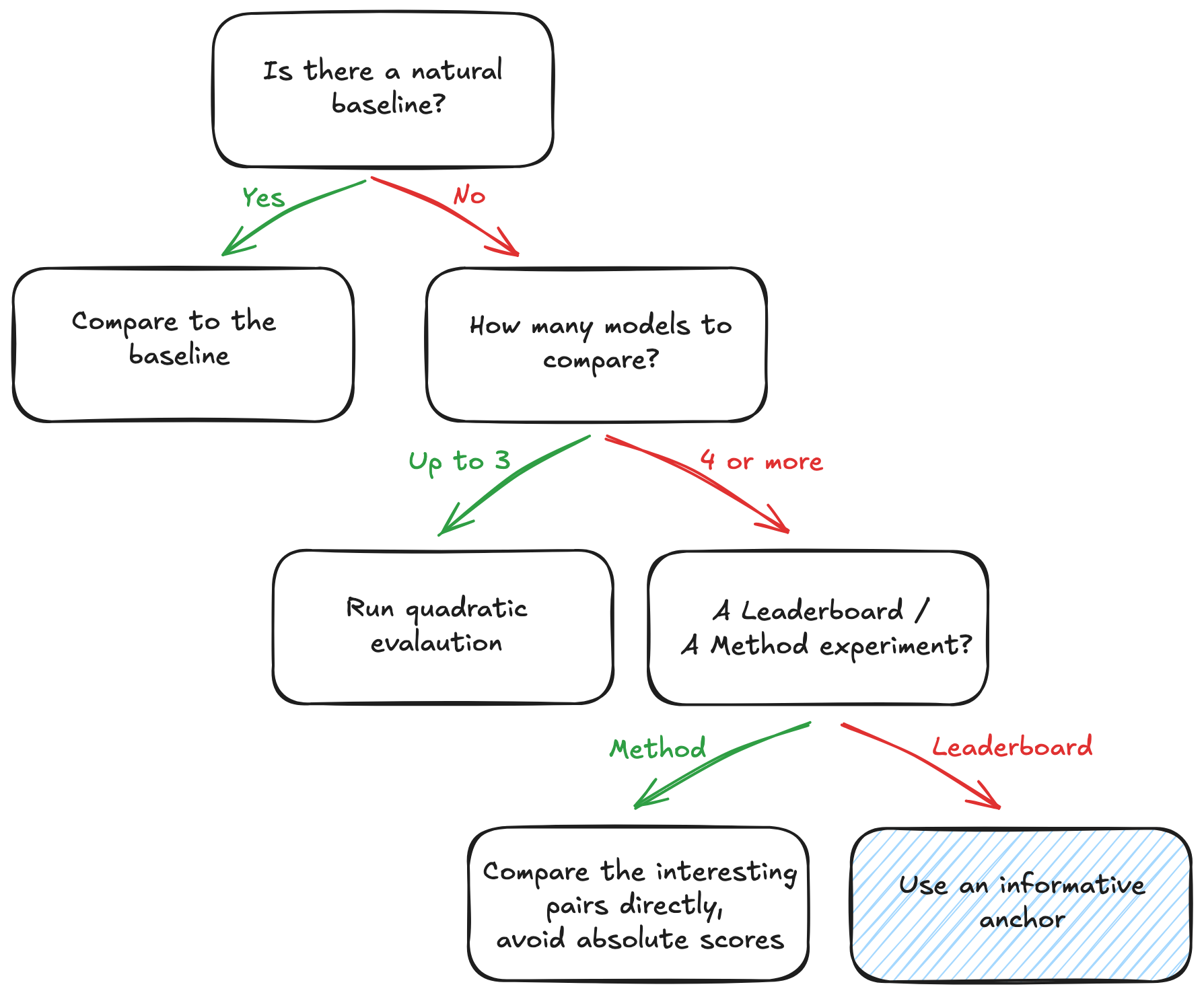}
    \caption{Decision tree for good pairwise evaluation.}
    \label{fig:tree}
\end{figure}

\section{Recommendations and Conclusion}

Based on our analysis, we propose good practices for LMJ anchor-based comparative evaluation.
Proposed recommendations are summarized in Fig.~\ref{fig:tree}. 

Our analysis revealed that a poor choice of anchor may throw away a substantial part of the evaluation budget, leading to noisier rankings.
We showed that a good anchor choice reduces the noise by up to $.19$ correlation points. However, even the best anchor we tested has $39\%$ of the benchmark's samples result in ties.
Hence, our first recommendation would be to \textbf{avoid anchor-based evaluation when possible}, as the pairwise setting is inherently limiting the informativeness\footnote{This could possibly be mitigated by a better judge who gives judgment on a broader scale, i.e. $v_{i,j}^{p,\mathcal{A}} \in \{-3, -2, -1, 0, 1, 2, 3\}$.} (\S\ref{sec:informative}). 

Passing through works that use anchor-based evaluation, we noticed a few cases of unnecessary use.
To eliminate such cases, we suggest first considering whether there is a natural anchor to which all the evaluated models should be compared. For example, a new training method is compared to a similar existing one. If so, use this newly trained model as the anchor, and avoid conclusions regarding other model pairs.

If there is no natural anchor and there are up to three models to evaluate, compare all pairs (quadratic evaluation). This will result in $3N$ judgments, the same as using an external anchor. 

If there are four or more models to evaluate, consider whether this is a leaderboard setting, i.e., you need to rank all the models. Sometimes, there are specific comparisons that are interesting, and there is no need to conclude regarding all the pairs. For example, a paper may propose a new method for a task, a `cheaper' version of it with a smaller model, and an enhancement to the method. We would like to compare the new method to some standard baseline of that task, the enhancement to the new method, and the cheap version to the baseline/new method, or maybe both of them. 
In this scenario, we will not report absolute scores, but rather the win-rates between pairs of interest.

When a full ranking of models is required, choose your anchor wisely. If possible, \textbf{use common knowledge, such as similar leaderboards, to avoid the strongest and weakest models}. Run your evaluation on a smaller sample set first, and confirm the informativeness of the chosen anchor (\S\ref{sec:estimate}). Finally, \textbf{report the anchor informativeness as part of your results} to reflect the validity of the evaluation.

\section{Related Work}

In the scope of this work, we discuss LMJ anchor-based pairwise evaluation (\S\ref{sec:intro}).
Another LMJ setting is pointwise evaluation, where, given an instruction and model response, the judge model provides an absolute quality score for the response. The score can be numeric, i.e., a number between 0 and 100, or categorical, e.g., [Very Bad, Bad, Mediocre, Good, Very Good] \citep{gu2024survey}.
Although pointwise evaluation is easier to scale, it has its own limitations. Its grading may be less suitable to differentiate between model pairs, and is less calibrated and robust to judge or prompt changes \citep{Evaluating_llmj}.

Many works investigated the effect of judge selection, demonstrating that stronger models generally align better with human preferences \citep{Evaluating_llmj, chatbot_arena, kocmi-federmann-2023-large}. Research has also extensively mapped systematic biases in judges, such as position bias \citep{wang-etal-2024}, verbosity bias \citep{saito2023verbosity}, and self-preference bias \citep{koo-etal-2024-benchmarking}.
However, while the choice of judge has been scrutinized, very few works have examined the anchor.
Current literature largely treats the anchor as a static default \citep{li2024crowdsourceddata, alpaca_eval}, overlooking its potential impact on the evaluation outcome. We show that anchor selection is crucial and equally important as the choice of the judge.

Few works have studied the effects of anchor-based evaluation. 
\citet{gao-etal-2025-evaluating} showed initial indications of a relation between the anchor performance and the evaluation quality. \citet{Non_Transitivity} demonstrated that LLM judges exhibit non-transitive preferences, leading to rankings that are sensitive to the choice of anchor. Similarly, \citet{wang2025trustjudge} highlighted discrepancies between pointwise and pairwise evaluations, as well as violations of transitivity. Both studies suggest alternative evaluation frameworks, such as dynamic matching strategies.
We, on the other hand, do not explore alternatives to the anchor-based evaluation. Instead, we identify that despite its drawbacks, it is extensively used, and propose practical improvements to this methodology.

Significant research has been dedicated to constructing robust benchmarks, with a particular focus on mitigating measurement artifacts such as saturation \citep{bowman-dahl-2021-will, ott2022mapping}. In light of this, we do not delve into these issues in this work, but instead proceed under the assumption that these established benchmarks are sufficiently representative of general capabilities and possess adequate discriminative power.

\section*{Limitations}

Our ``gold'' ranking is derived from the quadratic evaluation, which may correlate poorly with human rankings (see \S\ref{sec:vs_judges}). We therefore also report correlations with the human ranking.

We generated judgments with five different models, none of which is a commercial model. However, we chose top-performing open models and thus expect results to be applicable to top commercial models as well.
As for the evaluated models, we used both open and commercial models.


The standard BT method does not take into account the magnitude of the judgments. We therefore collapse the scores into $\{-1, 0, 1\}$ in experiments that use BT, losing information. In our power analysis, however, we do analyze both cases (with and without magnitude, i.e., Wilcoxon's signed rank test and the sign test).

``Mediocrity'' is defined strictly relative to the specific pool of evaluated models. Our findings indicate that an anchor is most effective when it is similar in capability to the models being compared, as this maximizes the discriminative signal. Consequently, if the evaluation focuses on a cluster of high-performing models, the optimal anchor must shift upwards to match them. Anchor selection, therefore, cannot be static; it must be continually calibrated to the capability range of the specific model set being ranked.

Anchor-based evaluation relies on the assumption of transitivity (if $A > Anchor$ and $Anchor > B$, then $A > B$), a property that LLM judges have been shown to violate at the instance level. However, we operate under the premise that while this assumption does not hold for every individual comparison, it remains sufficiently valid when averaged across the full benchmark. The aggregation of hundreds of pairwise verdicts helps mitigate the noise of specific intransitive cycles, yielding a global ranking that serves as a practical approximation of relative model quality.

\section*{Acknowledgements}
This research was partly supported by a grant from the Israel Science Foundation (grant no. 2912/25).

\bibliography{custom}

\appendix

\section{Full Correlation Results}
\label{app:full_results}
We evaluate the responses of $22$ contemporary models: \textit{o1}, \textit{o3 Mini}, \textit{GPT-4.1}, \textit{GPT-4.5 (Preview)}, \textit{o3 Mini High}, \textit{GPT-4.1 Mini}, \textit{o4 Mini}, \textit{GPT-4.1 Nano}, \textit{o3}, \textit{Qwen3 30B A3B}, \textit{Qwen3 32B}, \textit{QwQ 32B}, \textit{Qwen2.5 72B Instruct}, \textit{Qwen3 235B A22B} \citep{qwen3, qwen2_5}, \textit{Claude 3.7 Sonnet thinking 16k}, \textit{Claude 3.5 Sonnet}\footnote{\url{https://www.anthropic.com/news/claude-3-5-sonnet}}, \textit{Gemma 3 27B Instruct} \citep{gemmateam2025gemma3technicalreport}, \textit{Gemini 2.5 Flash} \citep{comanici2025gemini25pushingfrontier}, \textit{Llama 3.1 Nemotron 70B Instruct}, \textit{Llama 4 Maverick Instruct}\footnote{\url{https://ai.meta.com/blog/llama-4-multimodal-intelligence}}, \textit{Athene V2 Chat}, \textit{DeepSeek-R1} \citep{guo2025deepseek}.

\begin{figure*}[t]
\centering
\includegraphics[width=\textwidth]{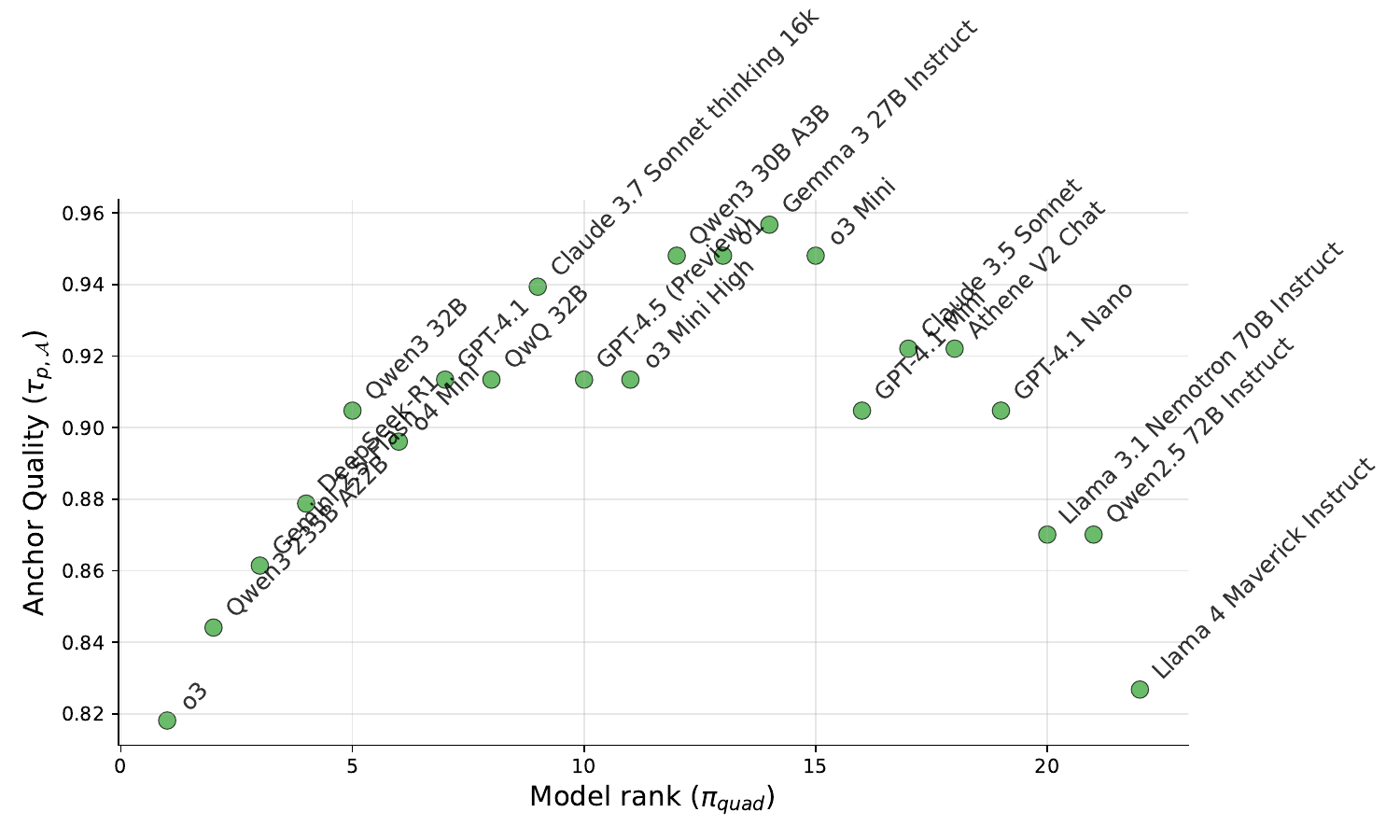}
\caption{Kendall’s $\tau$ correlation ($\tau_{p, \mathcal{A}}$) plotted against anchor position. The y-axis shows the correlation between the anchor-based ranking and the quadratic ranking $\pi_{quad}$, while the x-axis represents the anchor's position (rank) in $\pi_{quad}$. This reveals an inverted U-shaped relationship: top and bottom-ranked models correlate poorly with the gold standard, making them suboptimal anchors. The judge $J_p$ is \texttt{Deepeseek-V3}.}
\label{fig:arch_deepseek_full}
\end{figure*}

\begin{figure*}[t]
\centering
\includegraphics[width=\textwidth]{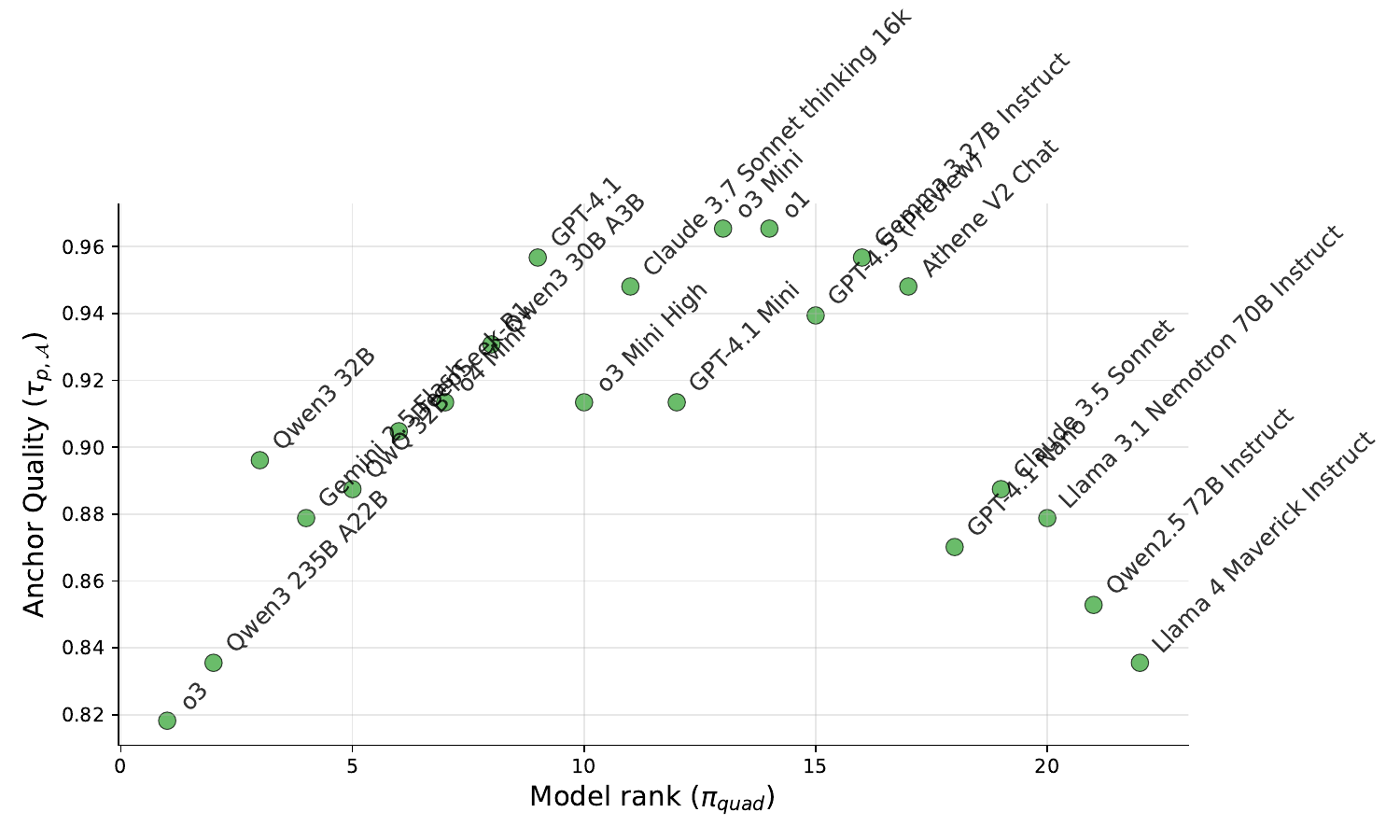}
\caption{Kendall’s $\tau$ correlation ($\tau_{p, \mathcal{A}}$) plotted against anchor position. The y-axis shows the correlation between the anchor-based ranking and the quadratic ranking $\pi_{quad}$, while the x-axis represents the anchor's position (rank) in $\pi_{quad}$. This reveals an inverted U-shaped relationship: top and bottom-ranked models correlate poorly with the gold standard, making them suboptimal anchors. The judge $J_p$ is \texttt{Qwen3 8B}.}
\label{fig:arch_small_qwen}
\end{figure*}

\begin{figure*}[t]
\centering
\includegraphics[width=\textwidth]{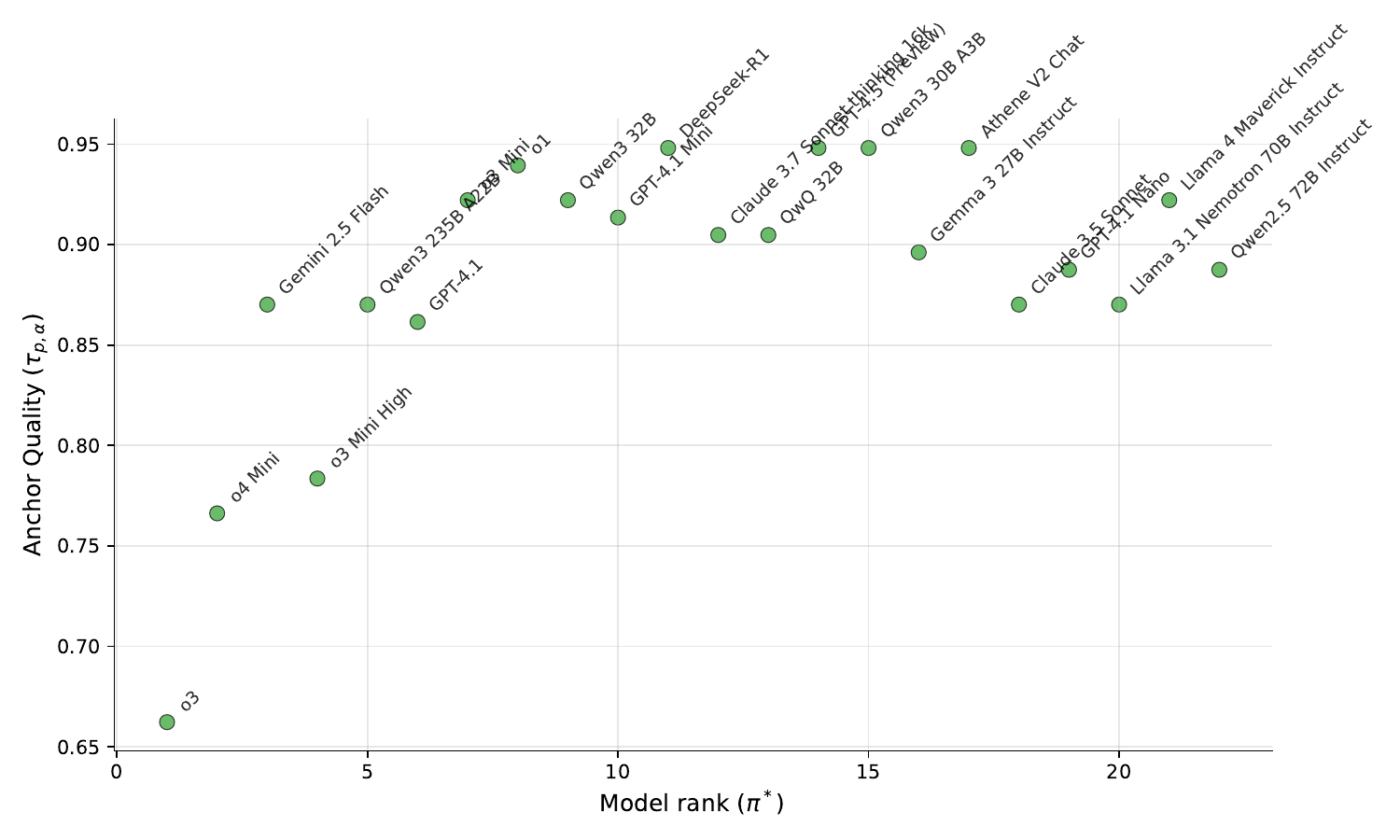}
\caption{Kendall’s $\tau$ correlation ($\tau_{p, \mathcal{A}}$) plotted against anchor position. The y-axis shows the correlation between the anchor-based ranking and the quadratic ranking $\pi_{quad}$, while the x-axis represents the anchor's position (rank) in $\pi_{quad}$. This reveals an inverted U-shaped relationship: top and bottom-ranked models correlate poorly with the gold standard, making them suboptimal anchors. The judge $J_p$ is \texttt{GPT-OSS 120B}.}
\label{fig:arch_large_gpt}
\end{figure*}

\begin{figure*}[t]
\centering
\includegraphics[width=\textwidth]{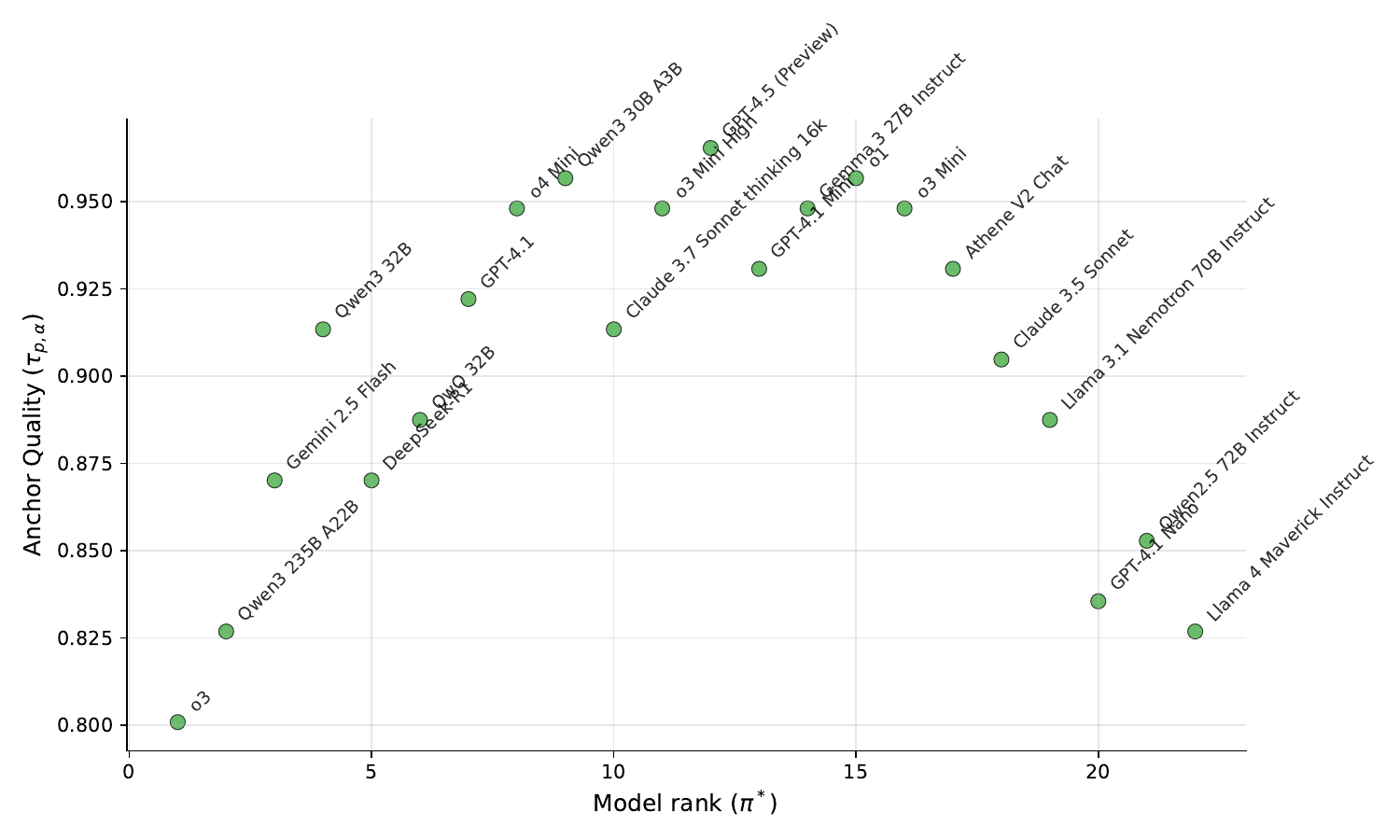}
\caption{Kendall’s $\tau$ correlation ($\tau_{p, \mathcal{A}}$) plotted against anchor position. The y-axis shows the correlation between the anchor-based ranking and the quadratic ranking $\pi_{quad}$, while the x-axis represents the anchor's position (rank) in $\pi_{quad}$. This reveals an inverted U-shaped relationship: top and bottom-ranked models correlate poorly with the gold standard, making them suboptimal anchors. The judge $J_p$ is \texttt{Qwen3 235B A22B}.}
\label{fig:arch_qwen}
\end{figure*}

\begin{figure*}[t]
\centering
\includegraphics[width=\textwidth]{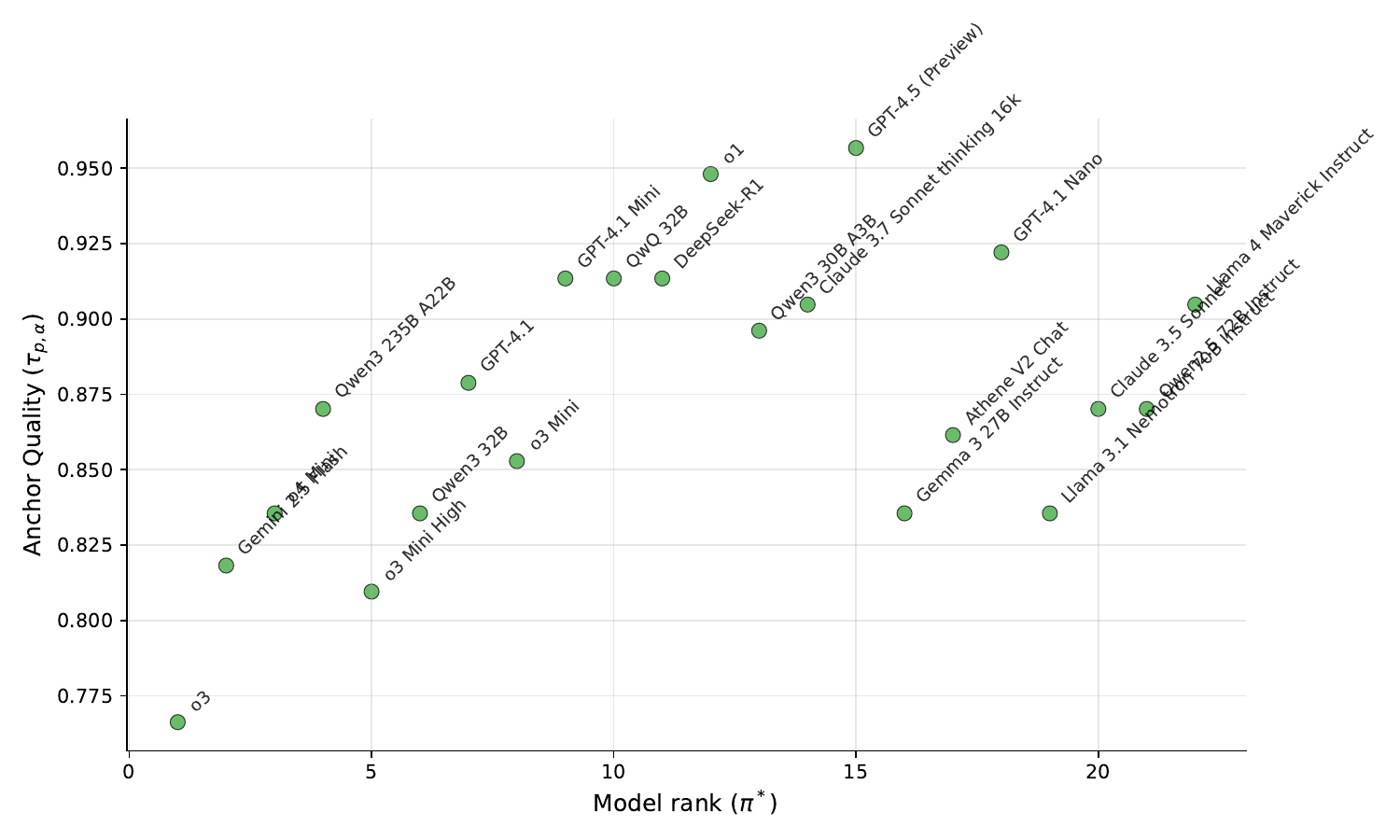}
\caption{Kendall’s $\tau$ correlation ($\tau_{p, \mathcal{A}}$) plotted against anchor position. The y-axis shows the correlation between the anchor-based ranking and the quadratic ranking $\pi_{quad}$, while the x-axis represents the anchor's position (rank) in $\pi_{quad}$. This reveals an inverted U-shaped relationship: top and bottom-ranked models correlate poorly with the gold standard, making them suboptimal anchors. The judge $J_p$ is \texttt{GPT-OSS 20B}.}
\label{fig:arch_small_gpt}
\end{figure*}

Here we provide the correlation results for all the judges (Tables.~\ref{tab:corr_small_qwen},~\ref{tab:corr_gpt_large},~\ref{tab:corr_qwen},~\ref{tab:corr_gpt_small}), and the correlations figures both with quadratic (Figs.~\ref{fig:arch_deepseek_full},~\ref{fig:arch_small_qwen},~\ref{fig:arch_large_gpt},~\ref{fig:arch_qwen},~\ref{fig:arch_small_gpt}), and human (Figs.~\ref{fig:arch_small_qwen},~\ref{fig:arch_large_gpt},~\ref{fig:arch_qwen},~\ref{fig:arch_small_gpt}) ranking. We can see that the inverted U shape trend remains, although the ranking is changing from judge to judge.
The human scores were acquired on September 11th, 2025.

\begin{figure*}[t]
\centering
\includegraphics[width=\textwidth]{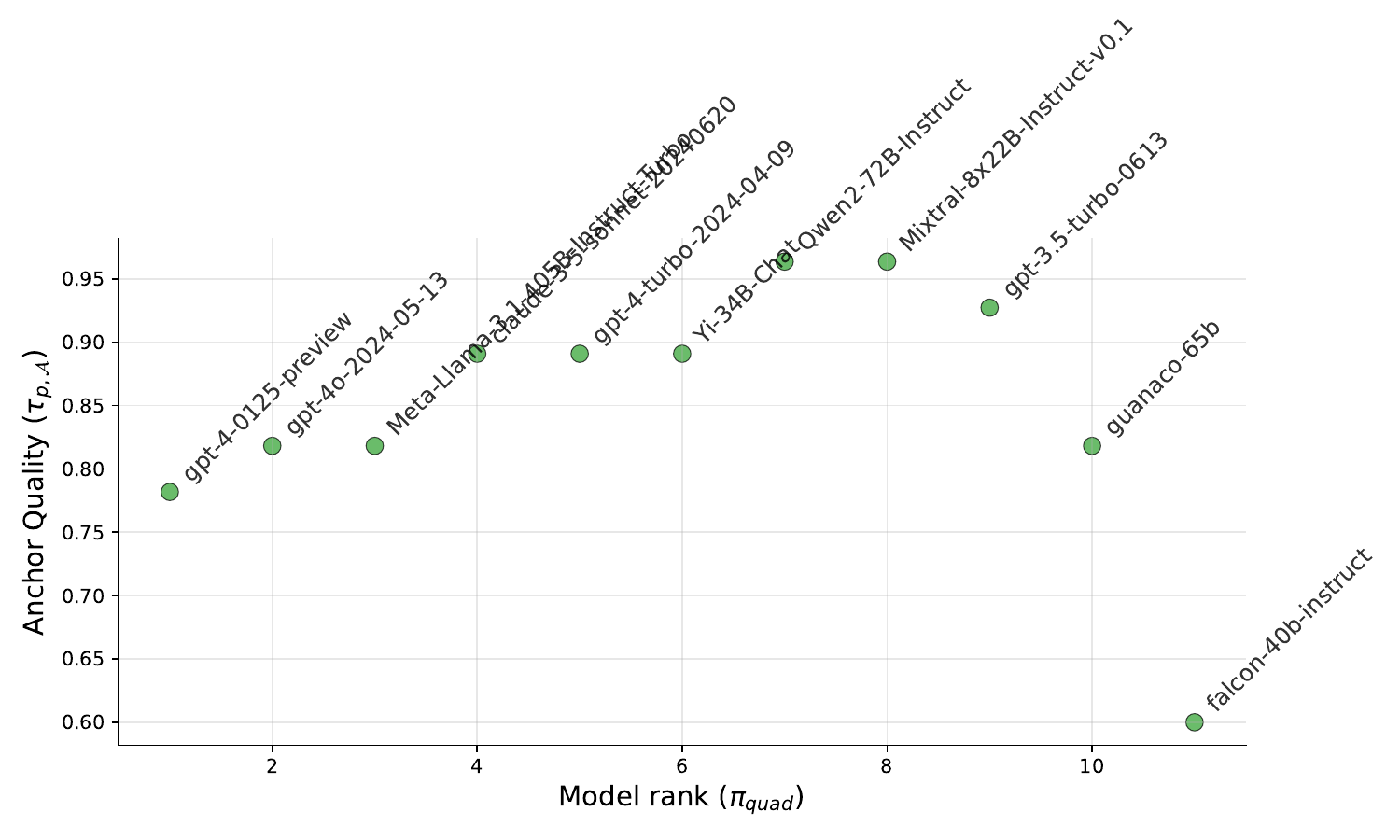}
\caption{Kendall's $\tau$ correlation, $\tau_{p, \mathcal{A}}$, of the anchor-based ranking with the quadratic ranking $\pi_{quad}$, plotted as a function of the anchor $m_\mathcal{A}$'s position in $\pi_{quad}$ on the AlpacaEval dataset. The judge $J_p$ is \texttt{DeepSeek-V3}. Top and bottom-ranked models in $\pi_{quad}$ correlate poorly with the quadratic ranking, making them suboptimal anchors.}
\label{fig:arch_alpaca_eval}
\end{figure*}

We replicate the results for the AlpacaEval dataset (\S\ref{sec:setup}), see Table.~\ref{tab:corr_alpaca} and Fig.~\ref{fig:arch_alpaca_eval}. We use $11$ models, which we chose based on their contemporaneity and performance: \textit{GPT-3.5 Turbo}, \textit{GPT-4 Turbo}, \textit{GPT-4o}, \textit{GPT-4 Turbo (Preview)}, \textit{Mixtral 8x22B Instruct} \citep{jiang2024mixtralexperts}, \textit{Qwen2 72B Instruct} \citep{qwen2}, \textit{Claude 3.5 Sonnet}, \textit{Llama 3.1 405B Instruct}, \textit{Yi 34B Chat} \citep{young2024yi}, \textit{Guanaco 65B} \citep{dettmers2023qlora}, \textit{Falcon 40B Instruct} \citep{almazrouei2023falconseriesopenlanguage}.

\begin{figure*}[t]
\centering
\includegraphics[width=\textwidth]{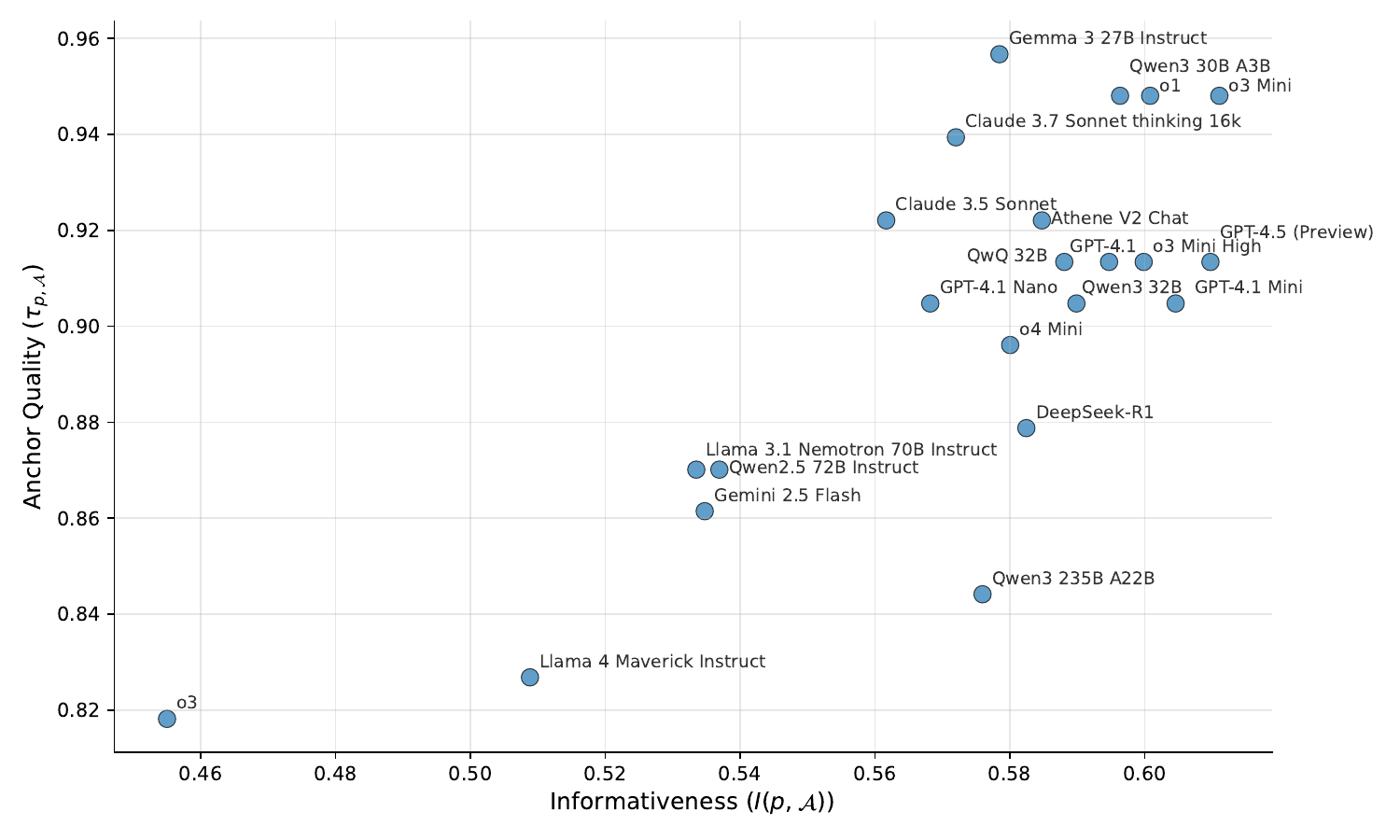}
\caption{Kendall’s $\tau$ correlation ($\tau_{p, \mathcal{A}}$) plotted against anchor informativeness. The y-axis shows the correlation between the anchor-based ranking and the quadratic ranking $\pi_{quad}$, while the x-axis represents the anchor's informativeness $I(p, \mathcal{A)}$. The plot exhibits a positive correlation between anchor quality and anchor informativeness. The judge is \textit{Deepseek-v3}.} \label{fig:informative_deepseek_vs_full}
\end{figure*}

\section{Informativeness}\label{app:informativeness}
Fig.~\ref{fig:informative_deepseek_vs_full} presents the plot from \S\ref{sec:informative} with all the labels.
Table.~\ref{tab:informative} show $I(p, \mathcal{A})$ for all anchors with \texttt{Deepseek-V3} as the judge. 

\section{Power Analysis Simulation}\label{app:power_analysis}
We provide the code for the power analysis simulation for the Wilcoxon signed test on our data, see \ref{lst:power_wide}.

We have a total of $22 \cdot \binom{21}{2} = 4620$ distributions, $143$ of them with effect size of $5\%$.

\begin{figure}[t]
    \centering
    \includegraphics[width=\columnwidth]{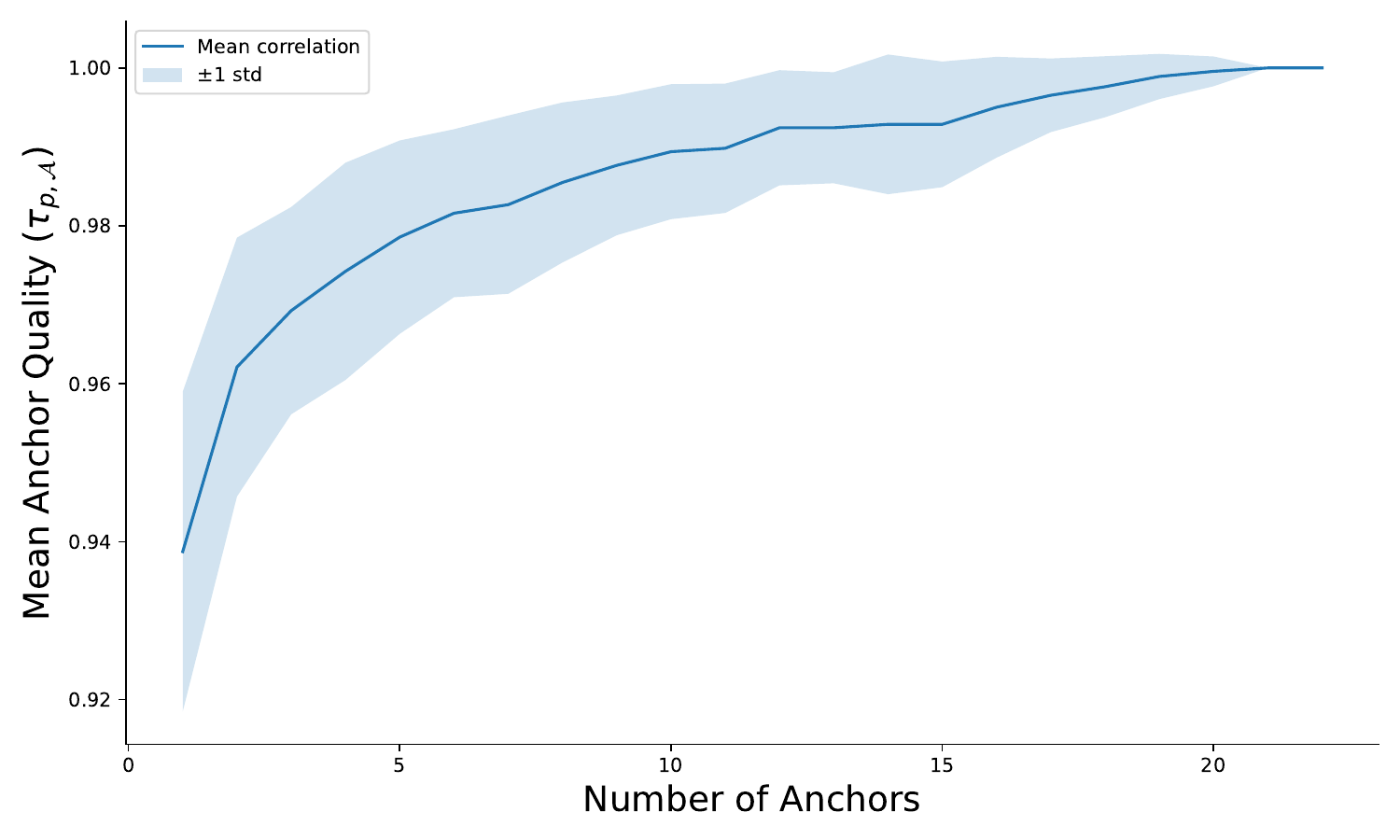}
    \caption{Mean $\tau_{p, \mathcal{A}}$ as a function of the number of anchors averaged over random anchor selections. 
    The correlation increases with the number of anchors, whereas the standard deviation decreases. However, this increase is smaller than the gap between choosing the strongest model as an anchor ($.82$) and a mean random choice ($.92$), demonstrating that while adding anchors helps, the initial anchor choice remains critical. The judge is \texttt{Deepseek-v3}.}
    \label{fig:num_anchors}
\end{figure}

\section{Number of Samples} \label{app:num_samples}
\begin{figure}[t]
    \centering
    \includegraphics[width=\columnwidth]{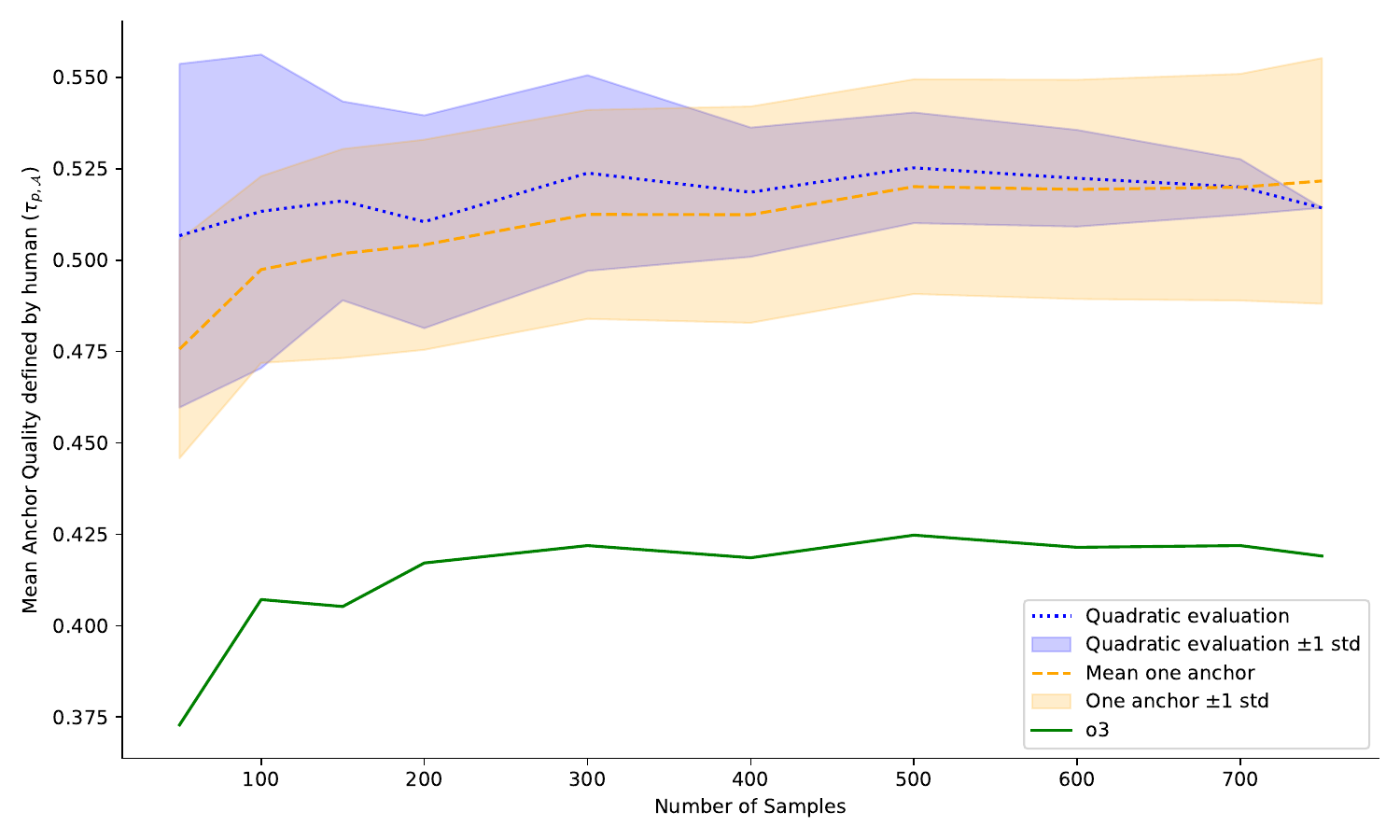}
    \caption{Mean $\tau_{p, \mathcal{A}}$ with respect to human ranking averaged over random sample selections as a function of sample size. 
    As the number of samples grows, the variance of the quadratic evaluation correlation decreases. Simultaneously, the mean anchor-based correlation improves, eventually converging with the quadratic correlation. This is not the case for each particular anchor choice, see \textit{o3} correlation. This demonstrates that anchor-based ranking is more affected by the dataset size than the quadratic ranking. The judge is \texttt{GPT-OSS 120B}.}
    \label{fig:num_samples_large_gpt}
\end{figure}

\begin{figure}[t]
    \centering
    \includegraphics[width=\columnwidth]{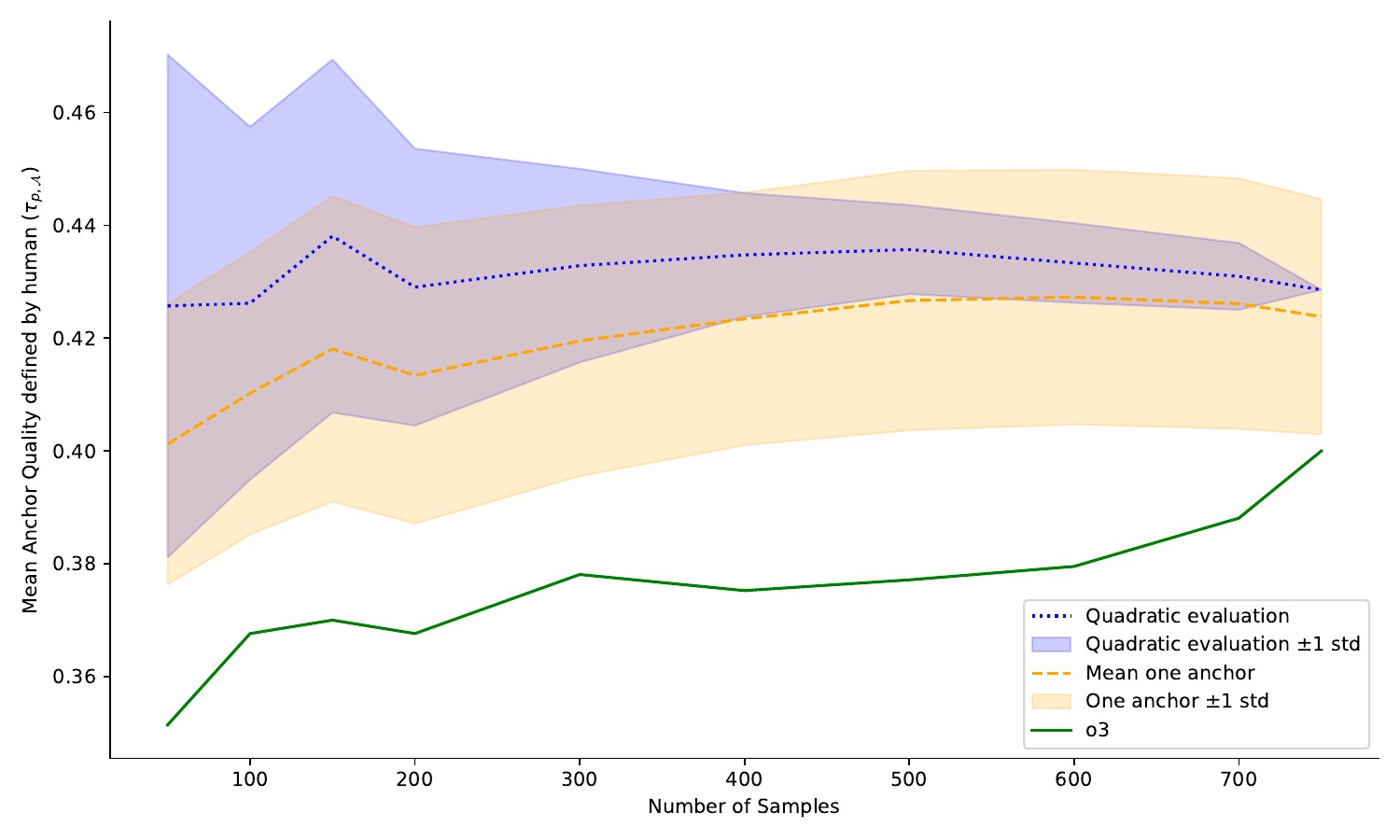}
    \caption{Mean $\tau_{p, \mathcal{A}}$ with respect to human ranking averaged over random sample selections as a function of sample size. 
    As the number of samples grows, the variance of the quadratic evaluation correlation decreases. Simultaneously, the mean anchor-based correlation improves, eventually converging with the quadratic correlation. This is not the case for each particular anchor choice, see \textit{o3} correlation. This demonstrates that anchor-based ranking is more affected by the dataset size than the quadratic ranking. The judge is \texttt{Qwen3 235B A22B Instruct}.}
    \label{fig:num_samples_small_gpt}
\end{figure}

\begin{figure}[t]
    \centering
    \includegraphics[width=\columnwidth]{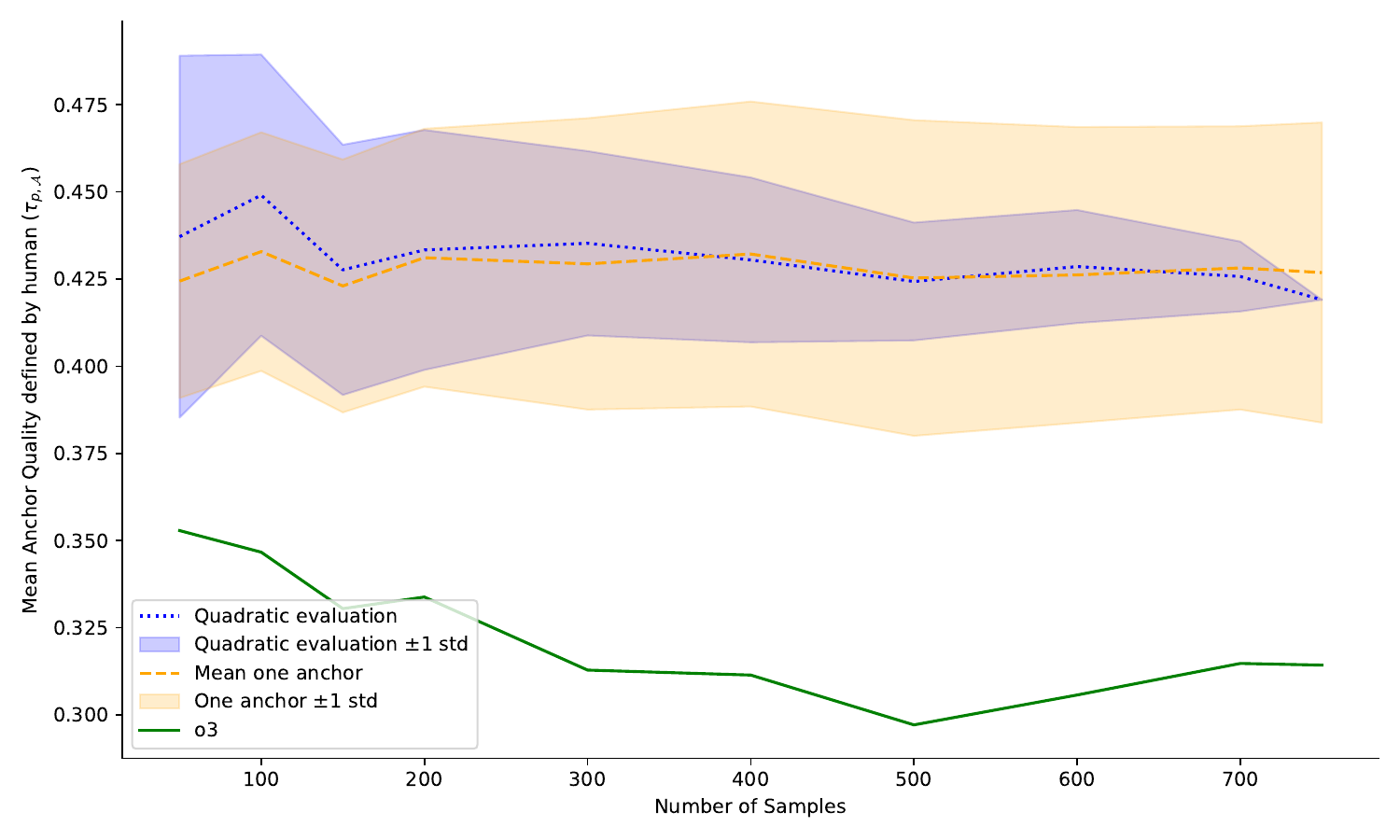}
    \caption{Mean $\tau_{p, \mathcal{A}}$ with respect to human ranking averaged over random sample selections as a function of sample size. 
    As the number of samples grows, the variance of the quadratic evaluation correlation decreases. Simultaneously, the mean anchor-based correlation improves, eventually converging with the quadratic correlation. This is not the case for each particular anchor choice, see \textit{o3} correlation. This demonstrates that anchor-based ranking is more affected by the dataset size than the quadratic ranking. The judge is \texttt{GPT-OSS 20B}.}
    \label{fig:num_samples_qwen}
\end{figure}

\begin{figure}[t]
    \centering
    \includegraphics[width=\columnwidth]{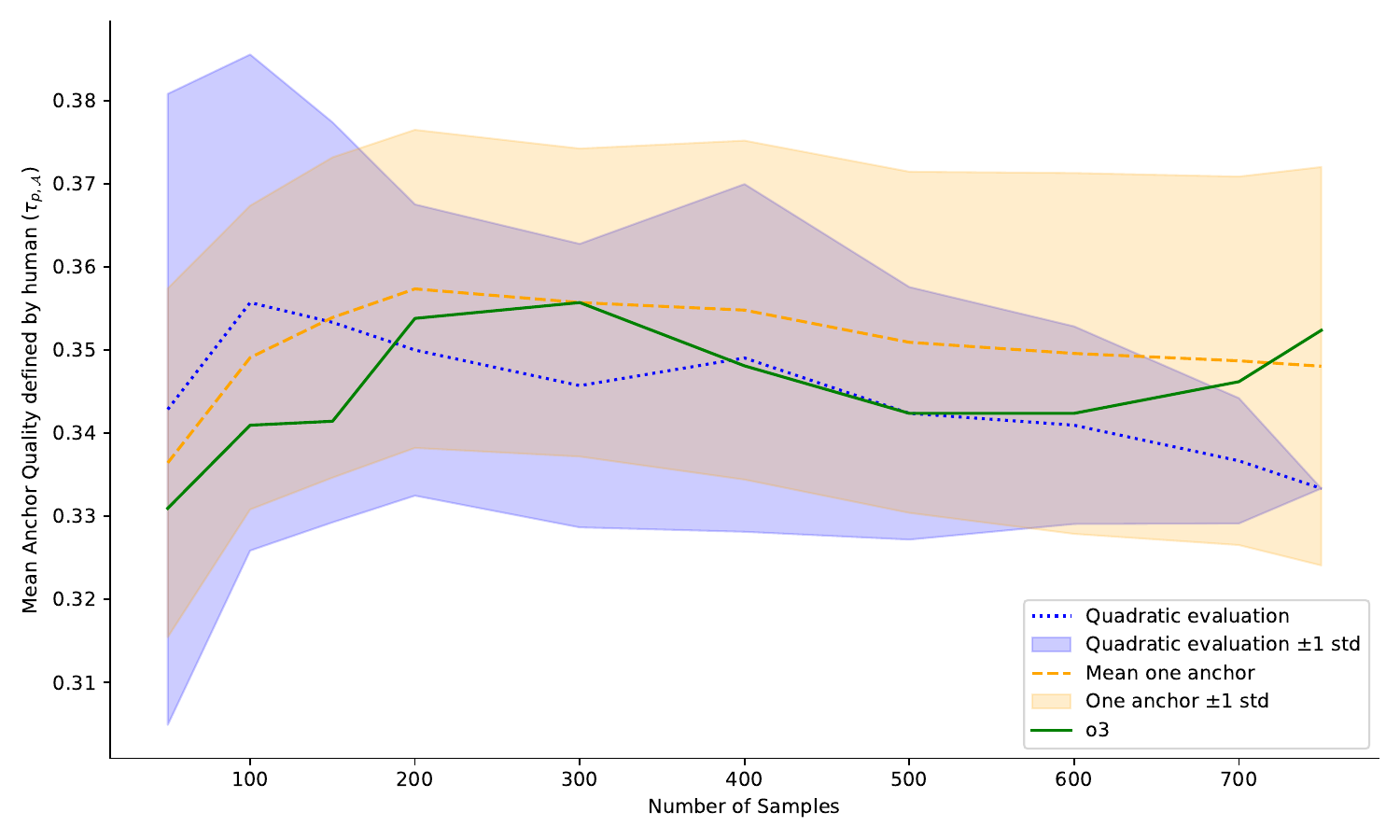}
    \caption{Mean $\tau_{p, \mathcal{A}}$ with respect to human ranking averaged over random sample selections as a function of sample size. 
    As the number of samples grows, the variance of the quadratic evaluation correlation decreases. 
    Unlike the case of large-medium judges, here with \texttt{Qwen3 8B} as the judge, the mean anchor-based correlation outperforms the quadratic correlation starting at approximately $150$ samples.}
    \label{fig:num_samples_small_qwen}
\end{figure}

We replicate the results from \S\ref{sec:data_size} for the other judges, measuring the effect of the dataset size on the anchor-based evaluation quality. 
We can see in Figs.~\ref{fig:num_samples_large_gpt},\ref{fig:num_samples_qwen}, \ref{fig:num_samples_small_gpt} that for the large-medium size judges, the main trends remain: the anchor-based evaluation is more affected by the size of the dataset than the quadratic evaluation. For the \texttt{Qwen3 8B} model, however, we see in Fig.~\ref{fig:num_samples_small_qwen} that the mean anchor-based correlation outperforms the quadratic correlation starting at approximately $150$ samples, and that the overall correlations are lower.

\section{Number of Anchors} \label{sec:multi_anchors}

Since relying on a single anchor introduces significant variance, we examine whether aggregating multiple anchors mitigates the issue. We perform an iterative analysis: starting with a single random anchor, we compute the Bradley-Terry ranking. In each subsequent step, we add another random model to the anchor set and recompute the ranking, continuing until all $22$ models are used (where the result converges to the quadratic ranking, correlation=$1.0$). We repeat this process over $40$ shuffled permutations and average their correlations $\tau_{p, \mathcal{A}}$ with the quadratic ranking. 

Fig.~\ref{fig:num_anchors} illustrates the mean correlation and standard deviation (shaded region) as a function of the anchor set size. As expected, correlation improves, and variance shrinks as more anchors are added. 
The mean correlation for a single random anchor is $0.92$, which seems high. Yet, as established in \S\ref{sec:intro}, practitioners do not select anchors at random; they typically select the strongest model (according to prior beliefs). Under this realistic constraint, the starting correlation is actually $0.82$ (see Table~\ref{tab:corr_combined})--a massive $.10$-point deficit compared to the random average. This demonstrates that while adding anchors helps, the initial choice of anchor remains a critical bottleneck for efficiency.

\section{Estimating Informativeness Full Results}\label{app:estimate}

The full results are provided in Table~\ref {tab:estimating_informativeness}. 

\begin{figure*}[t]
    \begin{lstlisting}[caption={Power analysis simulation.}, label={lst:power_wide}]
def run_power_analysis(all_N, effect_size, alpha, power, M):
    effect_size_interval = (effect_size, effect_size + 0.01)
    D = get_empirical_distributions_with_effect_size(effect_size_interval)
    
    for N in all_N:        
        null_hypothesis_rejected = 0
        for _ in range(M):
            d = np.random.choice(D)
            S = np.random.choice(d, N, replace=True)
            res = stats.wilcoxon(S, alternative='greater', zero_method='pratt')
            
            if res.pvalue < alpha:
                null_hypothesis_rejected += 1
        
        achieved_power = null_hypothesis_rejected / M
        if achieved_power >= power:
            print(f"Successfully rejected null hypothesis with N = {N}")
            break
    \end{lstlisting}
\end{figure*}

\begin{table}[t]
\centering
\small
\begin{tabular}{p{0.55\linewidth} c c}
\hline
\textbf{Anchor} & $\tau_{\text{quad}}$ & $\tau_{\text{human}}$ \\
\hline
o1 & .965 & .333 \\
o3 Mini & .965 & .352 \\
Gemma 3 27B Instruct & .957 & .305 \\
GPT-4.1 & .957 & .324 \\
Athene V2 Chat & .948 & .314 \\
Claude 3.7 Sonnet thinking 16k & .948 & .343 \\
GPT-4.5 (Preview) & .939 & .362 \\
Qwen3 30B A3B & .931 & .352 \\
o3 Mini High & .913 & .371 \\
GPT-4.1 Mini & .913 & .400 \\
o4 Mini & .913 & .314 \\
DeepSeek-R1 & .905 & .324 \\
Qwen3 32B & .896 & .410 \\
QwQ 32B & .887 & .381 \\
Claude 3.5 Sonnet & .887 & .381 \\
Gemini 2.5 Flash & .879 & .286 \\
Llama 3.1 Nemotron 70B Instruct & .879 & .362 \\
GPT-4.1 Nano & .870 & .352 \\
Qwen2.5 72B Instruct & .853 & .295 \\
Qwen3 235B A22B & .835 & .419 \\
Llama 4 Maverick Instruct & .835 & .305 \\
o3 & .818 & .257 \\
\hline
\textit{Average correlation:} & .904 & .343 \\
\textit{Standard deviation:} & .045 & .042 \\
\hline
\end{tabular}
\caption{Kendall's $\tau$ correlation, $\tau_{p, \mathcal{A}}$, of the anchor-based ranking with the quadratic and human ranking $\pi_{quad}$ and $\pi^{human}$ for \texttt{Qwen3-8B} as the judge.}\label{tab:corr_small_qwen}
\end{table}

\begin{table}[t]
\centering
\small
\begin{tabular}{p{0.55\linewidth} c c}
\hline
\textbf{Anchor} & $\tau_{\text{quad}}$ & $\tau_{\text{human}}$ \\
\hline
Athene V2 Chat & .948 & .533 \\
Qwen3 30B A3B & .948 & .533 \\
GPT-4.5 (Preview) & .948 & .533 \\
DeepSeek-R1 & .948 & .476 \\
o1 & .939 & .505 \\
Qwen3 32B & .922 & .543 \\
o3 Mini & .922 & .543 \\
Llama 4 Maverick Instruct & .922 & .533 \\
GPT-4.1 Mini & .913 & .533 \\
QwQ 32B & .905 & .543 \\
Claude 3.7 Sonnet thinking 16k & .905 & .505 \\
Qwen2.5 72B Instruct & .896 & .505 \\
Gemma 3 27B Instruct & .896 & .590 \\
GPT-4.1 Nano & .887 & .533 \\
Qwen3 235B A22B & .870 & .524 \\
Llama 3.1 Nemotron 70B Instruct & .870 & .552 \\
Claude 3.5 Sonnet & .870 & .581 \\
Gemini 2.5 Flash & .870 & .429 \\
GPT-4.1 & .861 & .438 \\
o3 Mini High & .784 & .533 \\
o4 Mini & .766 & .438 \\
o3 & .662 & .305 \\
\hline
\textit{Average correlation:} & .884 & .510 \\
\textit{Standard deviation:} & .069 & .062 \\
\hline
\end{tabular}
\caption{Kendall's $\tau$ correlation, $\tau_{p, \mathcal{A}}$, of the anchor-based ranking with the quadratic and human ranking $\pi_{quad}$ and $\pi^{human}$ for \texttt{GPT-OSS 120B} as the judge.}\label{tab:corr_gpt_large}
\end{table}

\begin{table}[t]
\centering
\small
\begin{tabular}{p{0.55\linewidth} c c}
\hline
\textbf{Anchor} & $\tau_{\text{quad}}$ & $\tau_{\text{human}}$ \\
\hline
GPT-4.5 (Preview) & .965 & .448 \\
Qwen3 30B A3B & .957 & .448 \\
o1 & .957 & .467 \\
Gemma 3 27B Instruct & .948 & .419 \\
o3 Mini High & .948 & .429 \\
o3 Mini & .948 & .429 \\
o4 Mini & .948 & .400 \\
Athene V2 Chat & .931 & .390 \\
GPT-4.1 Mini & .931 & .457 \\
GPT-4.1 & .922 & .381 \\
Claude 3.7 Sonnet thinking 16k & .913 & .400 \\
Qwen3 32B & .913 & .438 \\
Claude 3.5 Sonnet & .905 & .457 \\
QwQ 32B & .887 & .505 \\
Llama 3.1 Nemotron 70B Instruct & .887 & .390 \\
DeepSeek-R1 & .870 & .371 \\
Gemini 2.5 Flash & .870 & .352 \\
Qwen2.5 72B Instruct & .853 & .352 \\
GPT-4.1 Nano & .835 & .390 \\
Qwen3 235B A22B & .827 & .476 \\
Llama 4 Maverick Instruct & .827 & .429 \\
o3 & .801 & .295 \\
\hline
\textit{Average correlation:} & .902 & .415 \\
\textit{Standard deviation:} & .050 & .048 \\
\hline
\end{tabular}
\caption{Kendall's $\tau$ correlation, $\tau_{p, \mathcal{A}}$, of the anchor-based ranking with the quadratic and human ranking $\pi_{quad}$ and $\pi^{human}$ for \texttt{Qwen3 235B A22B Instruct} as the judge.}\label{tab:corr_qwen}
\end{table}

\begin{table}[t]
\centering
\small
\begin{tabular}{p{0.55\linewidth} c c}
\hline
\textbf{Anchor} & $\tau_{\text{quad}}$ & $\tau_{\text{human}}$ \\
\hline
GPT-4.5 (Preview) & .957 & .438 \\
o1 & .948 & .410 \\
GPT-4.1 Nano & .922 & .467 \\
DeepSeek-R1 & .913 & .429 \\
QwQ 32B & .913 & .457 \\
GPT-4.1 Mini & .913 & .448 \\
Claude 3.7 Sonnet thinking 16k & .905 & .381 \\
Llama 4 Maverick Instruct & .905 & .457 \\
Qwen3 30B A3B & .896 & .429 \\
GPT-4.1 & .879 & .324 \\
Claude 3.5 Sonnet & .870 & .457 \\
Qwen3 235B A22B & .870 & .476 \\
Qwen2.5 72B Instruct & .870 & .381 \\
Athene V2 Chat & .861 & .390 \\
o3 Mini & .853 & .419 \\
Qwen3 32B & .835 & .495 \\
Llama 3.1 Nemotron 70B Instruct & .835 & .343 \\
Gemma 3 27B Instruct & .835 & .495 \\
o4 Mini & .835 & .381 \\
Gemini 2.5 Flash & .818 & .390 \\
o3 Mini High & .810 & .419 \\
o3 & .766 & .200 \\
\hline
\textit{Average correlation:} & .873 & .413 \\
\textit{Standard deviation:} & .047 & .066 \\
\hline
\end{tabular}
\caption{Kendall's $\tau$ correlation, $\tau_{p, \mathcal{A}}$, of the anchor-based ranking with the quadratic and human ranking $\pi_{quad}$ and $\pi^{human}$ for texttt{GPT-OSS 20B} as the judge.}\label{tab:corr_gpt_small}
\end{table}

\begin{table}[t]
\centering
\small
\begin{tabular}{p{0.7\linewidth} c}
\hline
\textbf{Anchor} & $\tau_{\text{quad}}$ \\
\hline
Mixtral 8x22B Instruct & .964 \\
Qwen2 72B Instruct & .964 \\
GPT-3.5 Turbo & .927 \\
Claude 3.5 Sonnet & .891 \\
Yi 34B Chat & .891 \\
GPT-4 Turbo & .891 \\
Llama 3.1 405B Instruct & .818 \\
Guanaco 65B & .818 \\
GPT-4o & .818 \\
GPT-4 Turbo (Preview) & .782 \\
Falcon 40B Instruct & .600 \\
\hline
\textit{Average correlation:} & .851 \\
\textit{Standard deviation:} & .103 \\
\hline
\end{tabular}
\caption{Kendall's $\tau$ correlation, $\tau_{p, \mathcal{A}}$, of the anchor-based ranking with the quadratic ranking $\pi_{quad}$ for \texttt{Deepseek-V3} as the judge on the AlpacaEval dataset.} \label{tab:corr_alpaca}
\end{table}

\begin{figure*}[t]
\centering
\includegraphics[width=\textwidth]{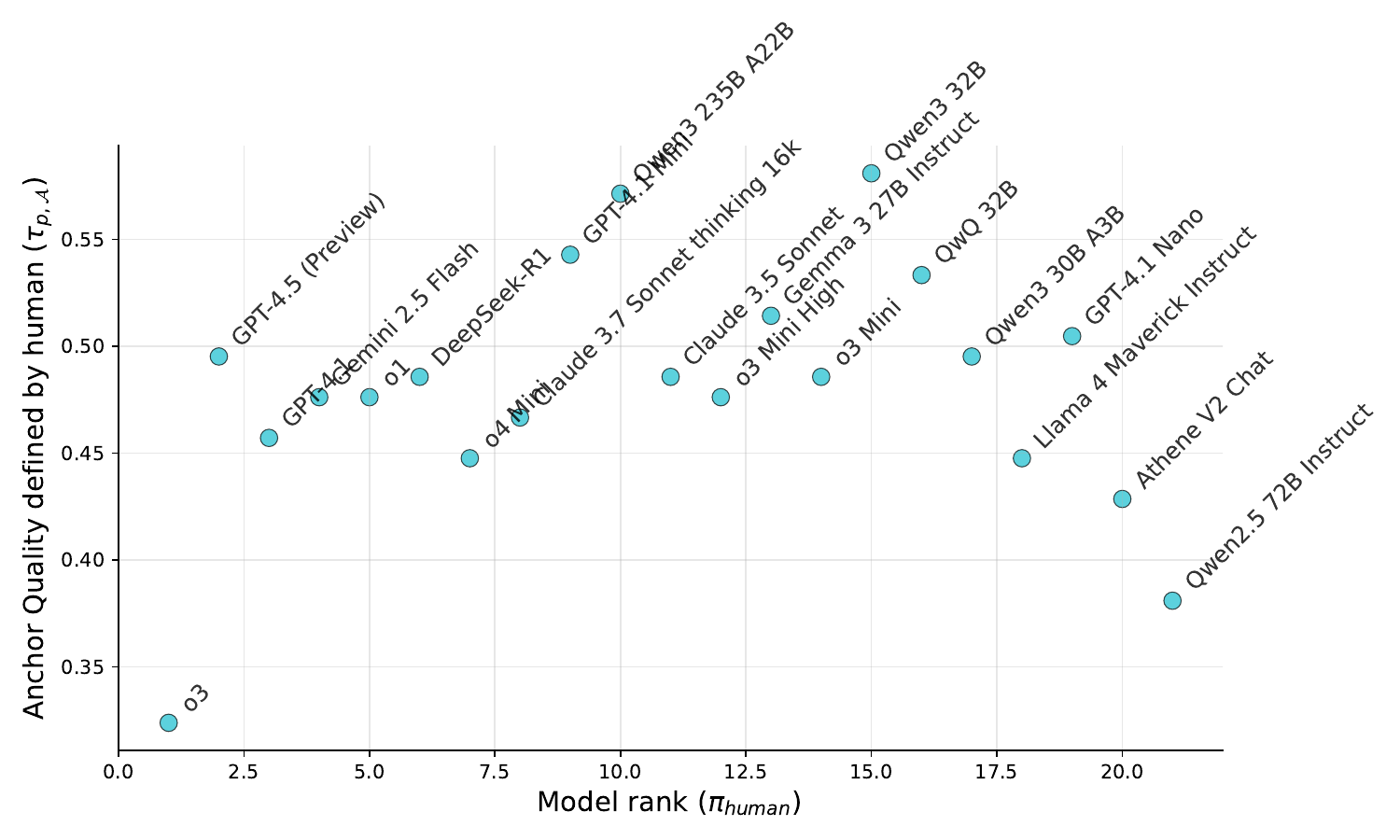}
\caption{Kendall's $\tau$ correlation, $\tau_{p, \mathcal{A}}$, of the anchor-based ranking with the human ranking $\pi_{human}$, plotted as a function of the anchor $m_\mathcal{A}$'s position in $\pi^{human}$. The judge $J_p$ is \texttt{DeepSeek-V3}. Top and bottom-ranked models in $_{human}$ correlate poorly with the human ranking, making them suboptimal anchors.}
\label{fig:arch_human}
\end{figure*}

\begin{figure*}[t]
\centering
\includegraphics[width=\textwidth]{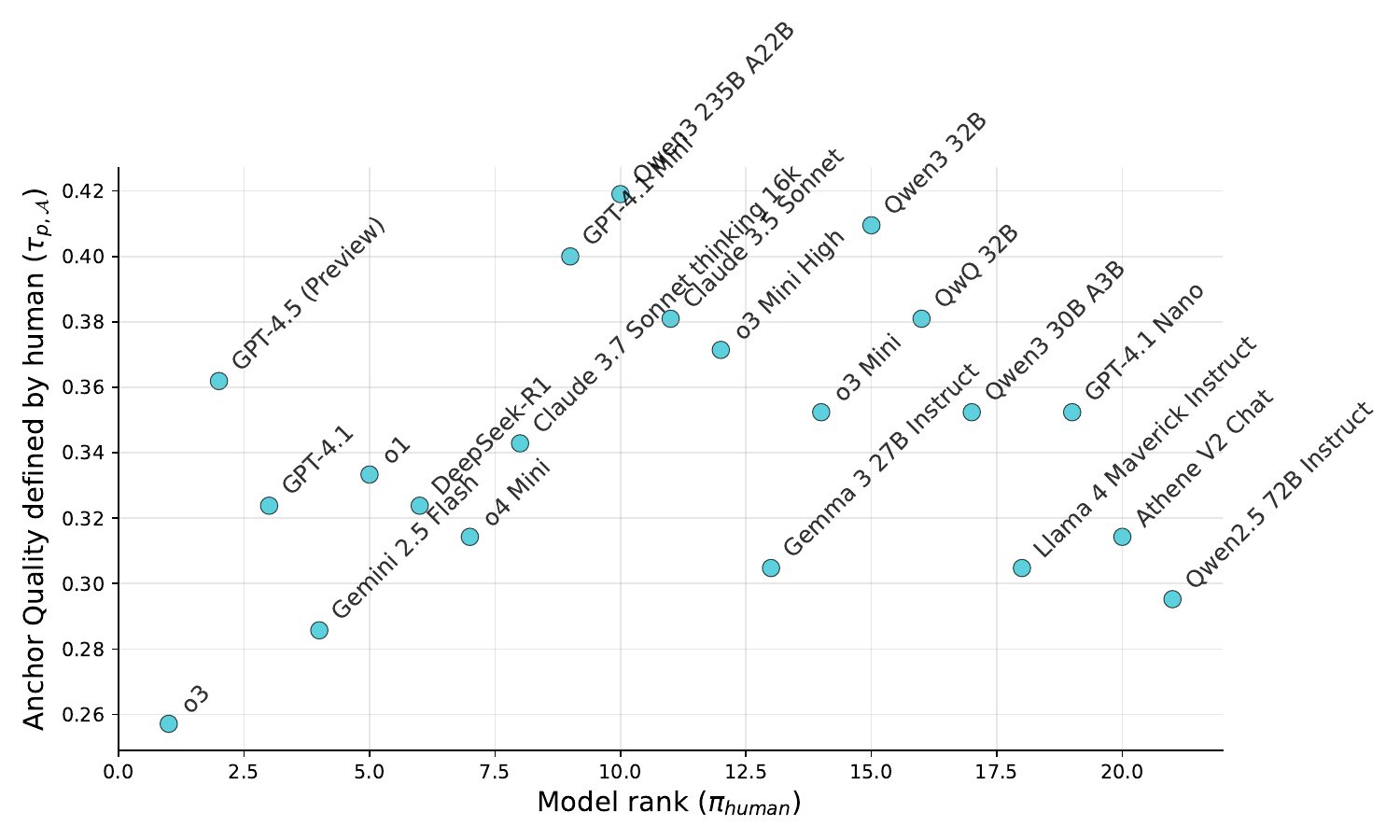}
\caption{Kendall's $\tau$ correlation, $\tau_{p, \mathcal{A}}$, of the anchor-based ranking with the human ranking $\pi_{human}$, plotted as a function of the anchor $m_\mathcal{A}$'s position in $\pi_{human}$. The judge $J_p$ is \texttt{Qwen3 8B}. Top and bottom-ranked models in $\pi_{human}$ correlate poorly with the human ranking, making them suboptimal anchors.}
\label{fig:arch_small_qwen_human}
\end{figure*}

\begin{figure*}[t]
\centering
\includegraphics[width=\textwidth]{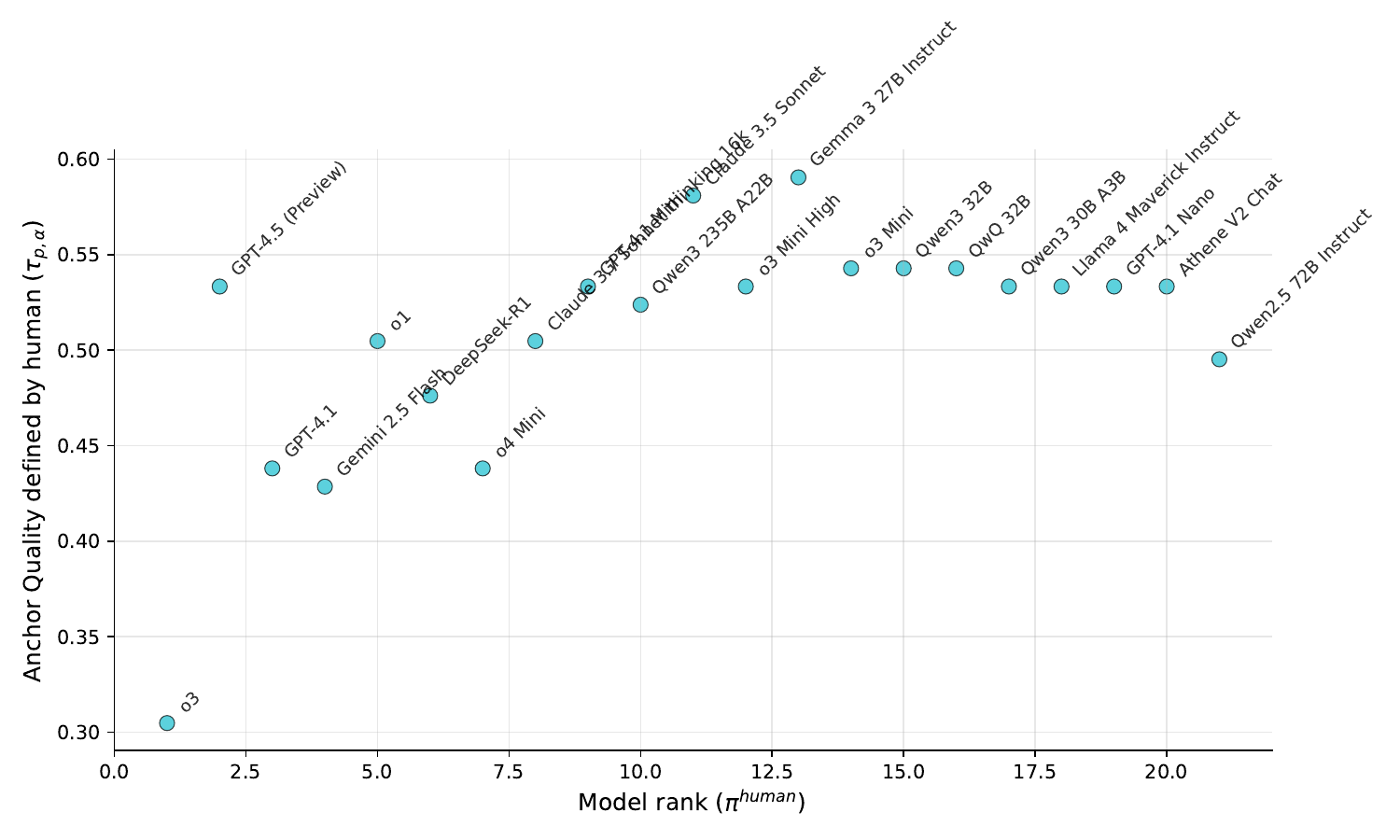}
\caption{Kendall's $\tau$ correlation, $\tau_{p, \mathcal{A}}$, of the anchor-based ranking with the human ranking $\pi_{human}$, plotted as a function of the anchor $m_\mathcal{A}$'s position in $\pi_{human}$. The judge $J_p$ is \texttt{GPT-OSS 120B}. Top and bottom-ranked models in $\pi_{human}$ correlate poorly with the human ranking, making them suboptimal anchors.}
\label{fig:arch_large_gpt_human}
\end{figure*}

\begin{figure*}[t]
\centering
\includegraphics[width=\textwidth]{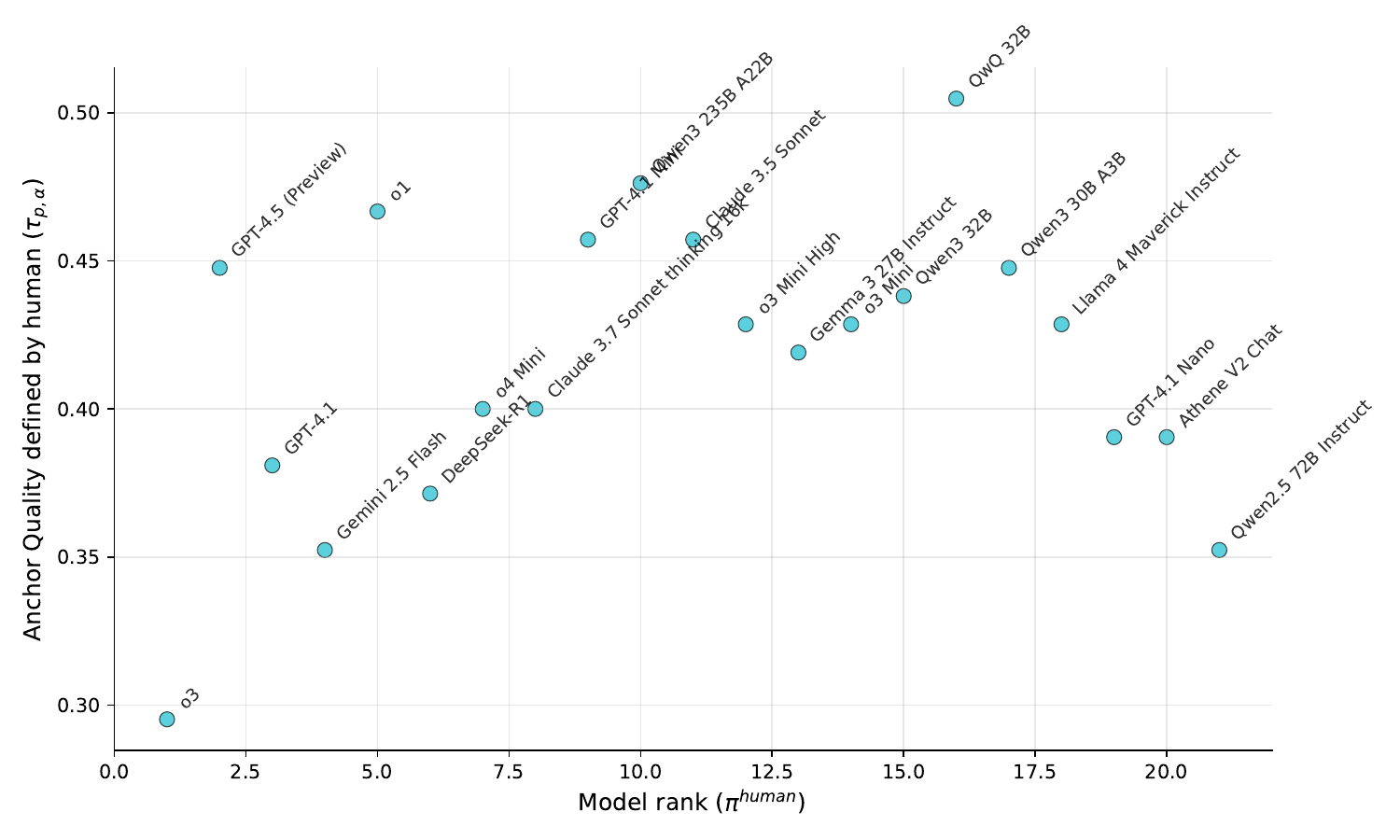}
\caption{Kendall's $\tau$ correlation, $\tau_{p, \mathcal{A}}$, of the anchor-based ranking with the human ranking $\pi_{human}$, plotted as a function of the anchor $m_\mathcal{A}$'s position in $\pi_{human}$. The judge $J_p$ is \texttt{Qwen3 235B A22B Instruct}. Top and bottom-ranked models in $\pi_{human}$ correlate poorly with the human ranking, making them suboptimal anchors.}
\label{fig:arch_qwen_human}
\end{figure*}

\begin{figure*}[t]
\centering
\includegraphics[width=\textwidth]{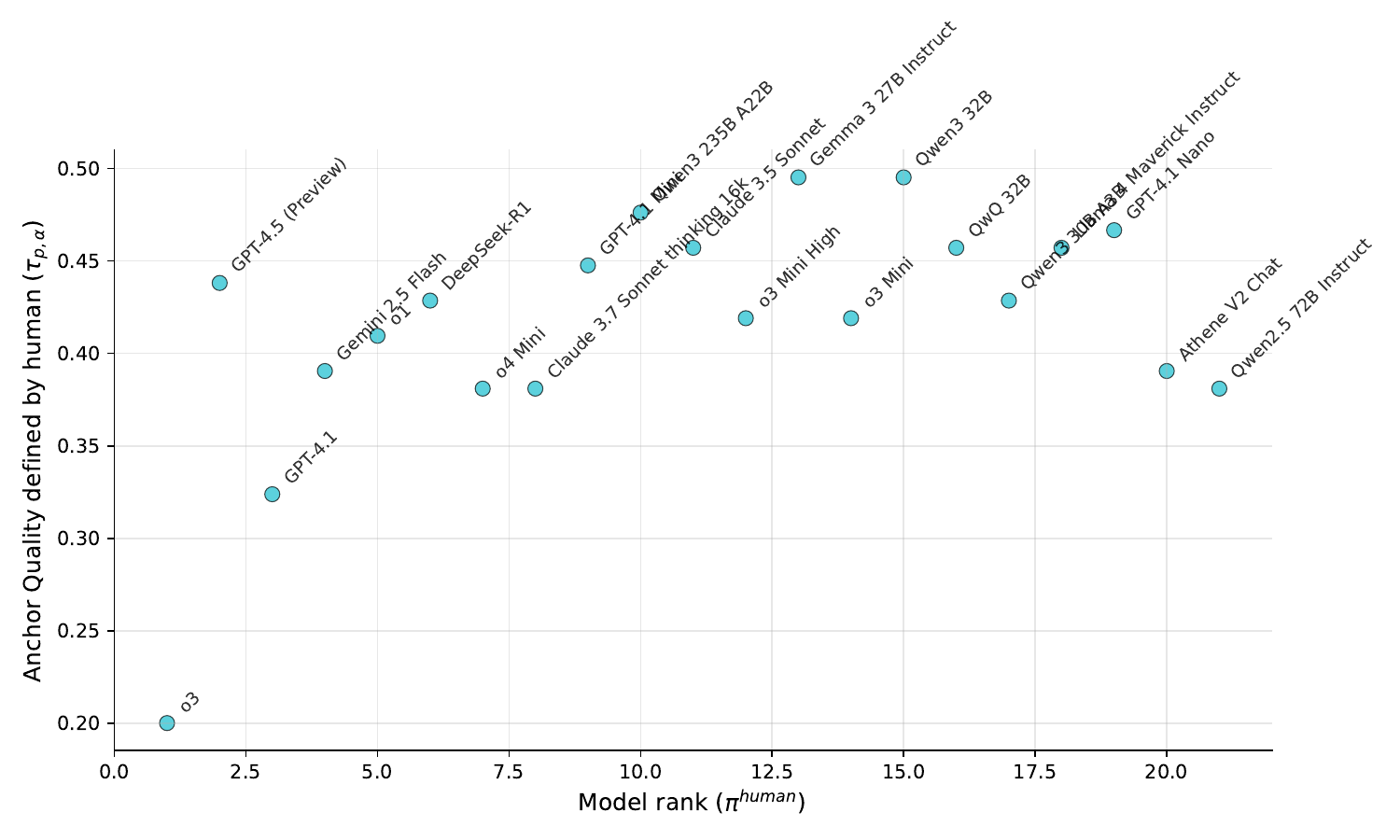}
\caption{Kendall's $\tau$ correlation, $\tau_{p, \mathcal{A}}$, of the anchor-based ranking with the human ranking $\pi_{human}$, plotted as a function of the anchor $m_\mathcal{A}$'s position in $\pi_{human}$. The judge $J_p$ is \texttt{GPT-OSS 20B}. Top and bottom-ranked models in $\pi_{human}$ correlate poorly with the human ranking, making them suboptimal anchors.}
\label{fig:arch_small_gpt_human}
\end{figure*}


\begin{table}[t]
\centering
\small
\setlength{\tabcolsep}{4pt}
\renewcommand{\arraystretch}{1.1}
\begin{tabular}{l r}
\toprule
\textbf{Model} & \textbf{Informative (\%)} \\
\midrule
o3 & 45.5 \\
Llama 4 Maverick Instruct & 50.9 \\
Llama 3.1 Nemotron 70B Instruct & 53.3 \\
Gemini 2.5 Flash & 53.5 \\
Qwen2.5 72B Instruct & 53.7 \\
Claude 3.5 Sonnet & 56.2 \\
GPT-4.1 Nano & 56.8 \\
Claude 3.7 Sonnet thinking 16k & 57.2 \\
Qwen3 235B A22B & 57.6 \\
Gemma 3 27B Instruct & 57.8 \\
o4 Mini & 58.0 \\
DeepSeek-R1 & 58.2 \\
Athene V2 Chat & 58.5 \\
QwQ 32B & 58.8 \\
Qwen3 32B & 59.0 \\
GPT-4.1 & 59.5 \\
Qwen3 30B A3B & 59.6 \\
o3 Mini High & 60.0 \\
o1 & 60.1 \\
GPT-4.1 Mini & 60.5 \\
GPT-4.5 (Preview) & 61.0 \\
o3 Mini & 61.1 \\
\bottomrule
\end{tabular}
\caption{Anchor informativeness with \texttt{Deepseek-V3} as the judge.}
\label{tab:informative}
\end{table}

\begin{table}[ht]
    \centering
    \begin{tabular}{cc}
        \toprule
        \textbf{Rank Pool Size} & \textbf{Pearson Correlation} \\
        \midrule
        3 Models  & 0.9090 \\
        4 Models  & 0.8652 \\
        5 Models  & 0.9528 \\
        6 Models  & 0.9367 \\
        7 Models  & 0.8909 \\
        8 Models  & 0.9658 \\
        9 Models  & 0.9578 \\
        10 Models & 0.9278 \\
        11 Models & 0.9500 \\
        12 Models & 0.9404 \\
        13 Models & 0.9438 \\
        14 Models & 0.9542 \\
        15 Models & 0.9471 \\
        16 Models & 0.9286 \\
        17 Models & 0.9679 \\
        18 Models & 0.9676 \\
        19 Models & 0.9518 \\
        20 Models & 0.9698 \\
        21 Models & 0.9100 \\
        22 Models & 0.9651 \\
        \bottomrule
    \end{tabular}
    \caption{Estimating anchor informativeness: Pearson Correlation of the estimated informativeness (with 10 samples) and the actual informativeness (with the full dataset) across different amounts of competitive models}
    \label{tab:estimating_informativeness}
\end{table}

\end{document}